\documentclass[10pt,twocolumn,letterpaper]{article}

\usepackage[pagenumbers]{cvpr} %

\usepackage{graphicx}
\usepackage{amsmath}
\usepackage{amssymb}
\usepackage{booktabs}
\usepackage{multirow}
\usepackage{paralist}
\usepackage{array}
\usepackage[normalem]{ulem}

\usepackage[pagebackref,breaklinks,colorlinks]{hyperref}

\usepackage[capitalize]{cleveref}
\crefname{section}{Sec.}{Secs.}
\Crefname{section}{Section}{Sections}
\Crefname{table}{Table}{Tables}
\crefname{table}{Tab.}{Tabs.}

\usepackage{verbatim}

\def\cvprPaperID{2489}  %
\def\confName{CVPR}
\def\confYear{2022}

\usepackage{dsfont}

\usepackage[usenames,dvipsnames]{xcolor}

\newcommand{\blockcomment}[1]{}

\newcommand{\netname}[0]{C3Det\xspace}

\usepackage{comment}

\newcommand{\late}[0]{LF\xspace}

\newcommand{\corr}[0]{C3\xspace}

\newcommand{\lossname}[0]{UEL\xspace}

\newcommand{\celldataset}[0]{LCell\xspace}
\newcommand{\smalldota}[0]{Tiny-DOTA\xspace}

\newcommand{\textbi}[1]{\ensuremath\textbf{\textit{#1}}}

\makeatletter
\renewcommand\paragraph{\@startsection{paragraph}{4}{\z@}%
                                    {1.00ex \@plus1ex \@minus.2ex}%
                                    {-1em}%
                                    {\normalfont\normalsize\bfseries}}

\makeatother

\begin{document}

\title{Interactive Multi-Class Tiny-Object Detection}

\author{
Chunggi Lee \hspace{3mm}
Seonwook Park \hspace{3mm}
Heon Song \hspace{3mm}
Jeongun Ryu \hspace{3mm}
\\
Sanghoon Kim \hspace{3mm}
Haejoon Kim \hspace{3mm}
S{\'e}rgio Pereira \hspace{3mm}
Donggeun Yoo \\[1mm]
Lunit Inc. \\[0.5mm]
{\tt\small \{cglee, spark, heon.song, rjw0205, seiker, oceanjoon, sergio, dgyoo\}@lunit.io
}
\vspace{-3mm}
}

\twocolumn[{%
\renewcommand\twocolumn[1][]{#1}%
\maketitle
\begin{center}
    \centering
    \captionsetup{type=figure}
    \vskip -4mm
    \includegraphics[width=\textwidth]{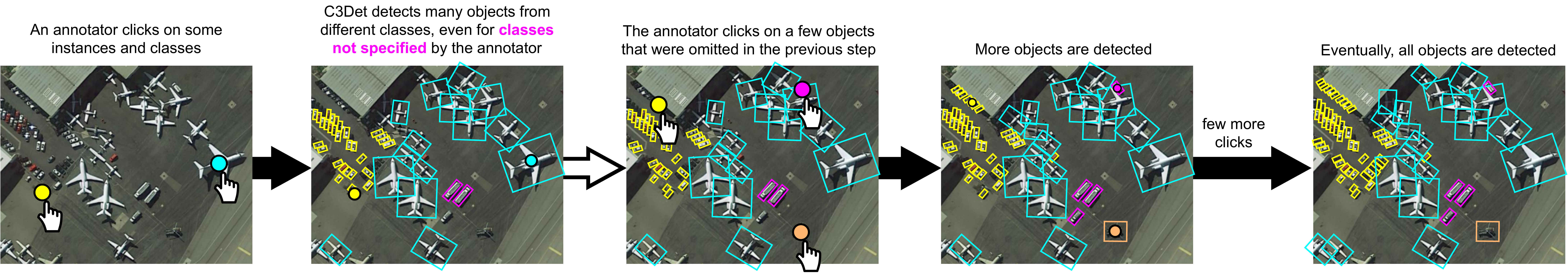}
    \vskip -3mm
    \captionof{figure}{
        \netname is a deep learning framework for interactive tiny-object detection that relates multiple annotator clicks to multiple instances and multiple classes of objects (each object class is depicted with a different color), in order to reduce overall annotation cost.
        \label{fig:teaser}
    }
    \vskip -1mm
\end{center}
}]

\begin{abstract}
Annotating tens or hundreds of tiny objects in a given image is laborious yet crucial for a multitude of Computer Vision tasks. 
Such imagery typically contains objects from various categories, yet the multi-class interactive annotation setting for the detection task has thus far been unexplored.
To address these needs, we propose a novel interactive annotation method for multiple instances of tiny objects from multiple classes, based on a few point-based user inputs.
Our approach, \netname, relates the full image context with annotator inputs in a local and global manner via late-fusion and feature-correlation, respectively.
We perform experiments on the \smalldota and \celldataset datasets using both two-stage and one-stage object detection architectures to verify the efficacy of our approach.
Our approach outperforms existing approaches in interactive annotation, achieving higher \textrm{mAP} with fewer clicks.
Furthermore, we validate the annotation efficiency of our approach in a user study where it is shown to be 2.85\texttt{x} faster and yield only 0.36\texttt{x} task load (NASA-TLX, lower is better) compared to manual annotation.
The code is available at \url{https://github.com/ChungYi347/Interactive-Multi-Class-Tiny-Object-Detection}.
\end{abstract}

\begin{figure*}[t]
\centering
\includegraphics[width=1.0\textwidth]{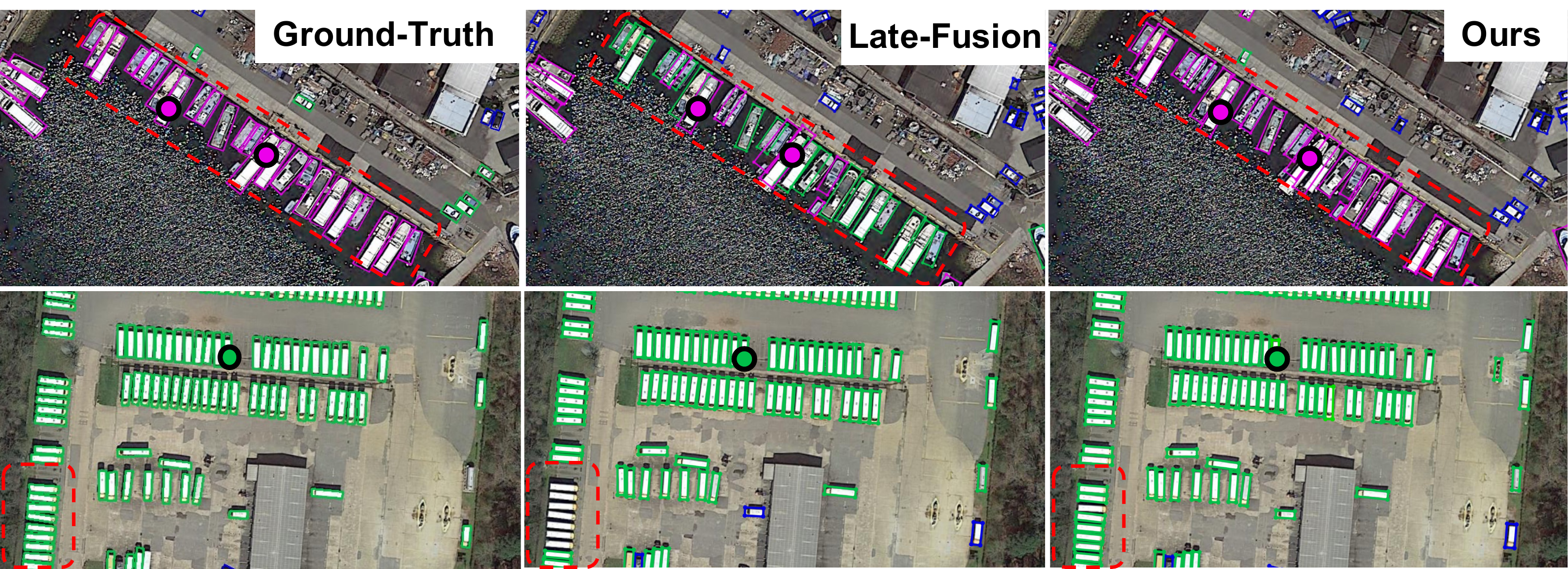}
\vskip -3mm
\caption{ 
    \textbf{The \textit{global context} of user inputs matter.} ``Late-Fusion'' does not consider the global context and can miss far away objects (red dotted lines) from user inputs (marked as circles). \netname captures the global context well and can detect far away objects.
    \label{fig:localglobal}
}
\vspace{-0.3cm}
\end{figure*}

\vspace{-3mm}
\section{Introduction}
\label{sec:intro}

Large-scale data and annotations are crucial for successful deep learning~\cite{mahajan2018exploring}.
However in many real-world problems, annotations are very labor-intensive and expensive to acquire~\cite{chui2018ai}.
Annotation costs increase even higher when handling numerous tiny objects such as in remote sensing \cite{cheng2016survey,li2020object,Xia_2018_CVPR}, extreme weather research \cite{racah2017extremeweather}, and microscope image analysis \cite{graham2019hover,li2018path}. 
These settings often require highly-skilled annotators and accordingly high compensation. 
For instance, cell annotation in Computational Pathology requires expert physicians (pathologists), whose training involves several years of clinical residency \cite{van2021deep,amgad2021nucls}. 
Reducing cost and effort for these annotators would directly enable the collection of new large-scale tiny-object datasets, and contribute to higher model performances.

Several prior works have been proposed to reduce annotation cost in other tasks.
Interactive segmentation methods~\cite{xu2016deep,maninis2018deep} focus on reducing the number of interactions in the segmentation of a \emph{single} foreground object, which can be classified as a ``many interactions to one instance'' approach.
However, tiny-objects annotation can benefit from a ``many interactions to many instances'' approach as one image can contain many instances.
Object counting methods~\cite{arteta2014interactive, ranjan2021learning} count multiple instances from a few user clicks and do follow a ``many interactions to many instances'' approach.
However, these methods highlight only objects of the \textit{same class} as the one being counted and thus can be classified as a ``one class to one class'' approach.
However, images with tiny-objects are often composed of objects from multiple classes.
Thus, tiny-object annotation should implement a ``many classes to many classes'' approach.

To address the above needs, we propose \netname, an effective interactive annotation framework for tiny object detection. \autoref{fig:teaser} shows how a user interacts with \netname to create bounding-boxes of numerous tiny objects from multiple classes. 
Once a user clicks on a few objects and provides their class information, \netname takes those as inputs and detects bounding boxes of many objects, even including object classes that the user did not specify. The user repeats this process until the annotation is complete. 
By utilizing user inputs in the ``many interactions to many instances'' and ``many classes to many classes'' way, \netname can significantly speed-up annotation.

A key aspect of our approach is in making each user click influence objects that are nearby (local context) as well as far away (global context).
To encourage the annotator-specified class to be consistent with model predictions, we insert user inputs (in heatmap form) at an intermediate stage in the model (late-fusion) and apply a class-consistency loss between user input and model predictions.
This alone can capture local context well, but may miss far away objects.
We therefore introduce the \corr (Class-wise Collated Correlation) module, a novel feature-correlation scheme that communicates local information to far away objects (see \autoref{fig:localglobal}), allowing us to learn many-to-many instance-wise relations while retaining class information.
Through extensive experiments, we show that these components combined, result in significant performance improvements. 

To validate whether our performance improvements translate to lower annotation cost in the real-world, we perform a user study with 10 human annotators.
Our approach, \netname, when combined with further manual bounding box corrections, is shown to be 2.85$\times$ faster and yield only 0.36$\times$ task load (NASA-TLX) compared to manual annotation, achieving the same or even better annotation quality as measured against the ground-truth.
This verifies that \netname not only shows improvements in simulated experiments, but also reduces annotation cost in the real-world.

In summary, we make the following contributions:
\begin{inparaenum}[(a)]
\item we address the problem of multi-class and multi-instance interactive annotation of tiny objects,
\item we introduce a training data synthesis and an evaluation procedure for this setting,
\item we propose a novel architecture for interactive tiny-object detection that considers both local and global implications of provided user inputs, and finally
\item our experimental results and user study verify that our method reduces annotation cost while achieving high annotation quality.
\end{inparaenum}

\begin{table*}[t]
    \centering
    \small
    \newcolumntype{M}[1]{>{\centering\arraybackslash}m{#1}}
    \begin{tabular}{|l|M{17mm}|M{21mm}|M{20mm}|M{30mm}|M{14mm}|}
        \hline
        Approach 
        & {\# of classes\newline to annotate} 
        & {Class-to-class relation} 
        & {\# of instances\newline to annotate} 
        & {Interaction-to-instance relation}
        & {Outputs} \\
        \hline
        Previous Interactive Detection~\cite{yao2012interactive}
        & 1 & 1-to-1 & many & many-to-many & bboxes \\
        Interactive Counting~\cite{arteta2014interactive, ranjan2021learning}
        & 1 & 1-to-1 & many & many-to-many & positions \\
        Interactive Segmentation~\cite{xu2016deep,maninis2018deep}        
        & 1 & 1-to-1 & 1 & many-to-1 & contour \\
        Ours                             
        & many & many-to-many & many & many-to-many & bboxes \\
        \hline
    \end{tabular}
    \vskip -2mm
    \caption{When considering the relationship between user interaction (typically the clicking of points) and annotated objects, we address the ``many interactions to many instances'' and ``many classes to many classes'' setting.
    }
    \label{tab:concept_comp}
\vspace{-3mm}
\end{table*}

\section{Related Work}

In this section, we discuss previous works that attempt to reduce annotation cost. The broad difference between our method and previous approaches are summarized in \autoref{tab:concept_comp}.

\paragraph{Interactive Object Detection}
The earliest work~\cite{yao2012interactive} in interactive detection incrementally trains Hough Forests over many image samples, gradually reducing false positives and negatives.
This method is shown to be effective in annotating cell images and pedestrian images, where there are typically no more than 20 instances from a single class. 
While~\cite{yao2012interactive} adopts incremental learning over 5 or more images, our CNN-based model can be applied immediately to new samples, and handle many more objects.

Closely related are weakly-supervised object detectors (WSOD) that take point input \cite{chen2021points, ren2020ufo} for establishing one-to-one correspondences between points and objects. 
However, this is different from our setting as these WSODs do not learn the many-to-many correspondences between points and objects, and do not yield interactive detectors.

\paragraph{Interactive Object Counting.}
This is similar to our tiny-object setting, since many tiny objects in an image are counted.
An early work~\cite{arteta2014interactive} learns a per-pixel ridge regression to adapt to user-provided point annotations, and counts instances of the specified object class.
More recently, in~\cite{ranjan2021learning}, a few box annotations of a target class are forwarded as user inputs to the counting model. 
However, these counting works consider the instances of a single object class, while \netname detects objects from multiple classes, including those not explicitly specified by user clicks.
Also, our method estimates accurate bounding boxes of all objects, while \cite{ranjan2021learning} outputs so-called density maps.

\paragraph{Interactive Object Segmentation.}
In this setting, users mark a few points on an image to yield a segmentation of a single foreground object. 
The earliest methods~\cite{boykov2001interactive} apply graph-cut, using an energy minimization method to separate foreground and background cues based on intensity changes, yielding optimal object boundaries.
Instead of requiring many exemplar strokes to indicate foreground versus background, GrabCut~\cite{rother2004grabcut} only requires a box to be drawn over the object.
Learning-based methods are first introduced in \cite{xu2016deep}, where users provide a few positive and negative clicks for object segmentation. The problem setting is formalized by introducing a training sample synthesis method, which is then followed by subsequent works \cite{li2018interactive,jang2019interactive,lin2020interactive,sofiiuk2020f,zhang2020interactive,ding2020phraseclick,acuna2018efficient,ling2019fast}.
\netname is similar to these learning-based methods in that we also synthesize user inputs from bounding box annotations for training but differs in the two following ways. 
First, for every user input, \netname annotates multiple objects from multiple classes at the same time, while these methods annotate a single object.
Second, \netname considers local-global relations to detect objects far away from the user's clicks, while interactive segmentation annotates just the object under the cursor.

\section{Overview}
Before we dive into describing the details of our method, we briefly motivate the broader decisions we made in building the \netname framework. 
All of our decisions are based on a simple (yet important) goal: reducing the real-world annotation cost of tiny-object detection.
This can be achieved by reducing annotation time and the number of interactions required from the annotator.
Our \netname framework addresses and improves on these aspects.

A quickly responding system improves both the annotation time and user experience. Taking inspiration from literature on deep interactive segmentation~\cite{xu2016deep}, we train a convolutional neural network (CNN) that only requires a simple feed-forward operation at test time. This has speed benefits compared to incremental learning approaches~\cite{yao2012interactive}.
We later show in~\autoref{sec:user_study} that our CNN-based system reduces annotation time significantly when compared with a fully manual annotation method, by operating at interactive rates\footnote{\netname responses takes just a few seconds on our user study GUI.}.

To further reduce the number of interactions required we devise two strategies. 
First, compared with bounding box inputs, we opt to receive the user inputs as point positions (via mouse click) along with object class.
This allows annotators to simply click on a tiny object and specify its class, yet yield full bounding boxes in the outputs of \netname.
Second, we decide to preemptively detect objects from classes that have not been selected by the annotator yet.
This allows the annotator to focus only on mistakes made by the annotation system.

Together with the contributions described in the following sections, we propose a meaningful solution to the problem setting of interactive multi-class tiny-object detection.

\section{Method}

In this section, we describe the proposed method. 
First, we introduce the overall architecture of \netname. Next, we describe a training data synthesis procedure for multi-class and multi-instance interactive object detection. Finally, we describe each component of \netname: the Late Fusion Module (\late), the Class-wise Collated Correlation Module (\corr), and User-input Enforcing Loss (UEL).

\subsection{Network Architecture}
\label{sec:network}

\netname detects objects in a given image guided by a few user inputs, and outputs bounding boxes and the class of as many objects as possible, including those that are not specified by such inputs.
We denote the input image as $\textbi{I}$, and the number of user inputs as $K$.

Each user input is denoted as $\left(\textbi{u}_k^{pos},\,u_k^{cls}\right)$, where $k$ is the index of user input, $\textbi{u}_k^{pos}$ defines a 2D position, and $u_k^{cls}\in\{1\ldots C\}$ is the object class. 
At inference time, $\left(\textbi{u}_k^{pos},\,u_k^{cls}\right)$ is provided by a user, while at training and validation time, it follows the center point and class of the chosen ground-truth bounding box.
Before passing the user inputs to the model, we convert each input $\left(\textbi{u}_k^{pos},\,u_k^{cls}\right)$ as a heatmap $\textbi{U}_k$ by placing a 2D Gaussian centered at $\textbi{u}_k^{pos}$ with a predefined standard deviation $\sigma_\mathrm{heatmap}$.

\begin{figure}[t]
    \centering
    \includegraphics[width=0.5\textwidth]{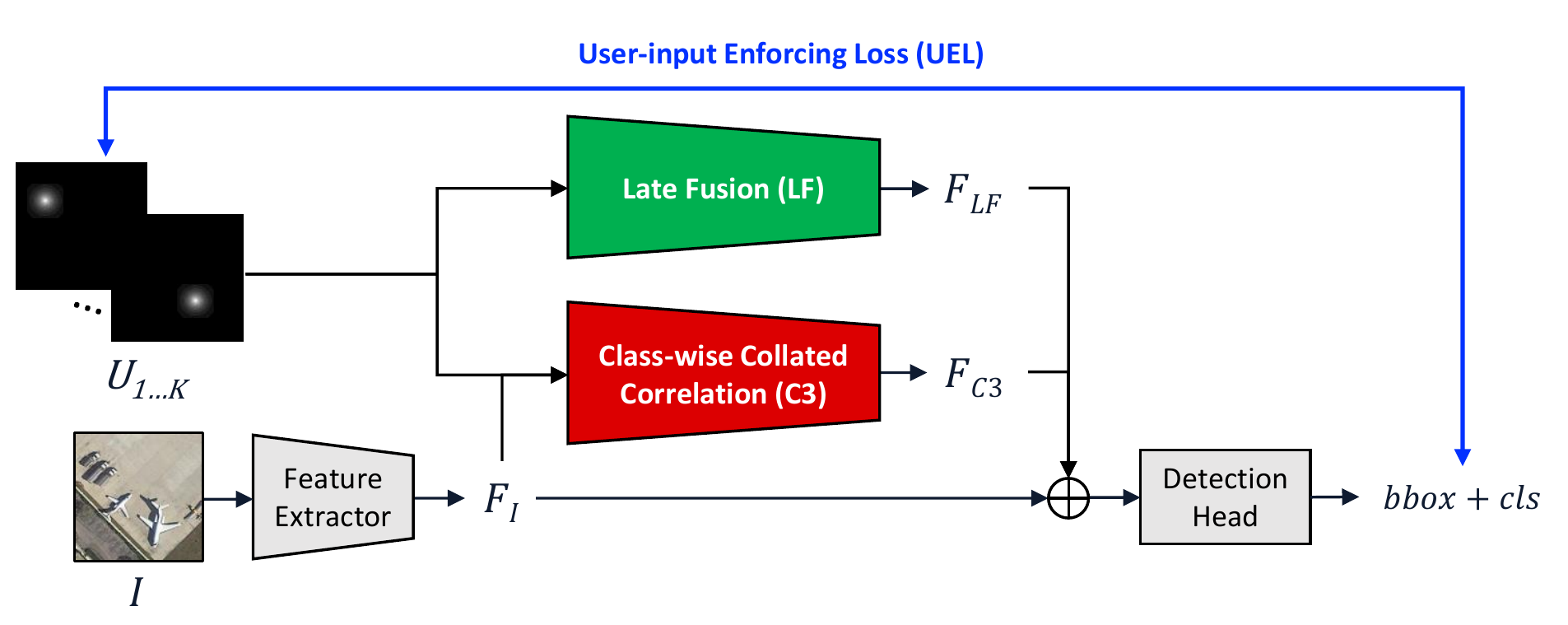}
    \vskip -3mm
    \caption{  
        \textbf{Overall architecture.} User inputs are processed and considered at both local (by Late Fusion) and global (by Class-wise Collated Correlation) context scales for multi-class multi-instance interactive tiny-object detection.
        The “$\oplus$” symbol indicates channel-wise concatenation. 
    }
    \label{fig_model_overview}
    \vspace{-0.3cm}
\end{figure}

The input image $\textbi{I}$ is first forwarded through a CNN feature extractor to yield a feature map $\textbi{F}_\mathrm{I}$.
Separately, the user input heatmaps, $\textbi{U}_{1\ldots K}$, are passed to the \late and \corr modules, which utilize user inputs in local and global manners, respectively.
The outputs of these modules, $\textbi{F}_\mathrm{LF}$ and $\textbi{F}_\mathrm{C3}$, are then concatenated to $\textbi{F}_\mathrm{I}$ before passing on to the next layers (see \autoref{fig_model_overview}). 

As \netname only modifies the outputs of the backbone network, it is applicable to both one-stage and two-stage architectures.
In the case of Faster R-CNN~\cite{ren2016faster} and RetinaNet~\cite{lin2017focal}, for example, the concatenated outputs are passed on to the region proposal network (RPN) and to the classification and box regression subnets, respectively.

\subsubsection{Training Data Synthesis}
\label{sec:training-simulation}

During training, we simulate the user inputs based on ground-truth annotations.
First, we randomly sample a target number of user inputs from a uniform distribution $N_u \sim \mathcal{U}_{[0, 20]}$.
While we define the uniform distribution to extend to $20$ only, this hyper-parameter can be adjusted as necessary.
We then sample $K = \mathrm{min}\left(N_u,\,N_a\right)$ objects (without replacement) from the ground-truth, where $N_a$ denotes the number of available objects for the current sample.
The object centers and class indices are then passed on to \netname as user inputs.

\subsubsection{Late Fusion Module (LF)}

When incorporating user input heatmaps to the network, two common approaches in interactive segmentation are early-fusion \cite{xu2016deep,xu2017deep,jang2019interactive,sofiiuk2020f} and late-fusion methods \cite{zhang2020interactive,agustsson2019interactive,rakelly2018few}.
Early-fusion methods concatenate user-input heatmaps to the input image, while late-fusion methods inject user-input heatmaps to an intermediate layer in the network, with \cite{zhang2020interactive} or without \cite{agustsson2019interactive,rakelly2018few} processing the heatmaps with CNN layers.
Prior insights show that late-fusion outperforms early-fusion \cite{rakelly2018few,zhang2020interactive}, and we find that this is also the case for interactive tiny-object detection.

To handle a varying number of user inputs, while maintaining the class information of the given inputs, we group the $K$ user input heatmaps by class, then apply a pixel-wise $\mathrm{max}$ operation to each group to yield $C$ heatmaps.
For the case where no inputs are provided for an object class, we simply pass a heatmap filled with zeros.
The heatmaps are passed to the \late module (a CNN-based feature extractor such as ResNet-18) that outputs feature maps $\textbi{F}_\mathrm{LF}$.

The \late module handles these heatmaps without any global pooling, and therefore does not lose any spatial information.
For the local area around a user input $\textbi{u}_k^{pos}$, the predicted objects' class can be directly affected by the user input $u_k^{cls}$.
We can therefore consider the \late module as one that considers the \emph{local context} of user inputs.

\subsubsection{Class-wise Collated Correlation Module (\corr)}

\begin{figure}[t]
\centering
\includegraphics[width=1.0\columnwidth]{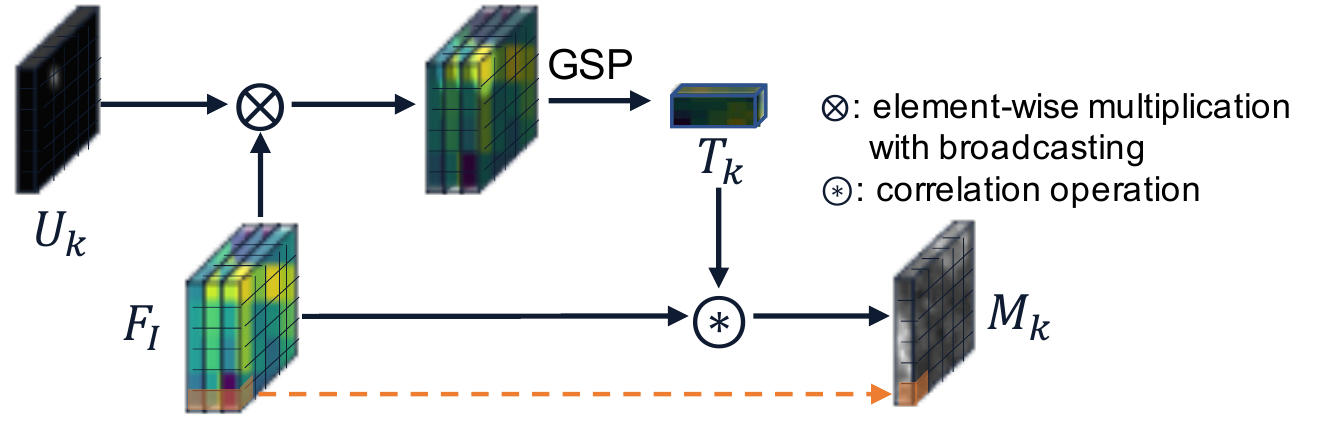}
\vskip -3mm
\caption{ 
    \textbf{Procedure of generating a correlation map.} A template vector is extracted from a feature map based on a user-input. The correlation map is computed between the template vector and the feature map. GSP stands for ``global sum pooling''.
    \label{fig:correlation}
}
\vspace{-0.3cm}
\end{figure}

While understanding the local context can help in predicting the correct class for objects near to user inputs, objects far away from user inputs must be impacted in a different way.
Recently, in \cite{ranjan2021learning}, a correlation operation between $\textbi{F}_\mathrm{I}$ and user input related features was used to improve object counting performance, using a few exemplars to count as many similar objects as possible in a given image.
Similarly, we suggest to extract template features from $\textbi{F}_\mathrm{I}$ based on user inputs, perform correlation with $\textbi{F}_\mathrm{I}$ (see \autoref{fig:correlation}), and merge the correlation maps class-wise.

For each provided user input heatmap $\textbi{U}_k$\footnote{The heatmap is typically resized to match the size of $\textbi{F}_\mathrm{I}$ and normalized such that it sums to 1.}, we perform the following to obtain a ``template'' vector,
\begin{equation}
    \textbi{T}_{k}\left(i\right) = \sum_{x,y} 
    \textbi{F}_{\mathrm{I}}\left(i,x,y\right)\, 
    \textbi{U}_k\left(x,y\right),
\end{equation}
where $i$ refers to a channel index and $x$, $y$ the column and row indices in $\textbi{F}_\mathrm{I}$ and $\textbi{U}_k$. This template vector can then be used as follows to generate a correlation map $\textbi{M}_k$,
\begin{equation}
    \textbi{M}_{k}\left(x,y\right) = \sum_{i} 
    \textbi{T}_{k}\left(i\right)\,
    \textbi{F}_\mathrm{I}\left(i,x,y\right).
\end{equation}

Once $K$ correlation maps are computed, we combine them class-wise based on $u_k^{cls}$, via an element-wise $\mathrm{max}$ operation as defined by,
\begin{equation}
    {{\textbi{F}_\mathrm{C3}}\left(c,x,y\right)=\max\{\textbi{M}_{k}\left(x,y\right)| u_k^{cls} = c, \forall{k} \in \left[K\right]\}},
\end{equation}
where $c$ refers to a class index.
Classes that do not have any associated user inputs are simply represented by a correlation map filled with zeros. 

This reduction allows us to produce $C$ correlation maps to pass on to the next stages, no matter how many user inputs are provided.
We describe this approach as a \emph{correlate-then-collate} method, where each user input is handled independently.
An intuitive alternative is a \emph{collate-then-correlate} method, where user input heatmaps are combined by class first, to perform the correlation operation once per object class.
The \emph{collate-then-correlate} alternative may be more robust to the choice of user input, but also assumes that each object class can be described by a single feature representation.
In our ablation study (see \autoref{ablation_corr_accum}), we show that the \emph{correlate-then-collate} method performs better, and thus choose this to define our \corr module.

The explicit correlation operation performed by the \corr module allows for local features to be compared across the entire image.
This extends the effect of user inputs on the model's predictions beyond the considerations of the \late module. 
In other words, we can consider the \corr module as one that considers the \emph{global context} of user inputs by learning many-to-many instance-wise relations.

\subsubsection{User-input Enforcing Loss (UEL)}

When a user specifies an object to be of a certain class, \netname should reflect this class on its prediction. 
Therefore, we propose to apply a training-time consistency loss between user inputs and model predictions through a User-input Enforcing Loss that enforces a class-wise consistency.

For each simulated user input, $(\textbi{u}_k^{pos},\,u_k^{cls})$, we retrieve the associated ground-truth bounding box $\textbi{y}_k^{bbox}$. 
We compare each of these ground-truth objects with all $J$ predicted objects (indexed by $j\in\{1\ldots J\}$).
Each prediction consists of a bounding box $\hat{\textbi{y}}_j^{bbox}$ and class $\hat{y}^{cls}_j$.
To compute the loss, we check for a non-zero intersection-over-union  ($IoU$) between every input-prediction pair, and apply a class-consistency loss.
The full loss is formulated as,
\begin{equation}
\mathcal{L}_{\mathrm{UEL}} =
\sum_{j,\, k}
\mathds{1}_{IoU\left(
    \hat{\textbi{y}}^{bbox}_j,
    \textbi{y}^{bbox}_k
\right)\,>\,0}
\,
\cdot
\,
\ell(\hat{y}_j^{cls}, u_k^{cls})
  \label{eq:uel}
\end{equation}
where $\ell$ is a loss function such as the cross entropy loss or the focal loss, depending on the main task loss.

\section{Experimental Results}

To validate the proposed approach, we train and evaluate on two multi-class tiny-object datasets, \smalldota and \celldataset.
We compare the performance of \netname against several baseline methods, and show that \netname applies to both one-stage and two-stage detectors such as RetinaNet and Faster R-CNN, respectively.
These are standard baselines architectures for detecting oriented bounding boxes on the DOTA dataset~\cite{Xia_2018_CVPR, wang2020tiny}, and our method should apply to other object detection architectures as well\footnote{See supplementary materials for results on RoI Transformer\cite{ding2019learning}.}.
Furthermore, we present ablation studies to verify the efficacy of our modules. 
For the implementation details of our model, please refer to our supplementary materials.

\subsection{Datasets}

\begin{table}
\small
\centering
\setlength{\tabcolsep}{0.2em}
\begin{tabular}{|l|c|c|c|c|c|c|c|}
\hline
\multirow{2}{*}{} & \multirow{2}{8.8mm}{Num. classes} & \multicolumn{3}{c|}{Num. patches}                           & \multicolumn{3}{c|}{Mean objects / patch} \\ \cline{3-8} 
             &   & train & val  & test & train & val & test  \\ \hline
\smalldota   & 8 & 11198 & 1692 & 2823 & 38.5 & 42.8 & 35.3 \\ \hline
\celldataset & 8 & 3681  & 250  & 823  & 79.3 & 82.0 & 99.6 \\ \hline
\end{tabular}
\vskip -2mm
\caption{
Comparison of statistics between \smalldota and \celldataset.
The number of patches reported for \smalldota are counted after subdividing the original images as described in \autoref{section_smalldota}.
\celldataset contains a higher mean number of objects per patch.
\label{tab:datasets}}
\end{table}

\paragraph{\smalldota} 
\label{section_smalldota}

The DOTA dataset~\cite{Xia_2018_CVPR,ding2021object} consists of aerial images and includes a variety of tiny (e.g. vehicles and ships) and larger objects (e.g. soccer ball field and basketball court). 
Following~\cite{wang2020tiny} which deals with tiny object detection, we filter out the larger objects in the DOTA v2.0 dataset, yielding 8 tiny-object classes. 
Furthermore, our procedure requires frequent querying of the test-set ground-truth (for validation-time user-input synthesis), yet the original DOTA test set's labels are not publicly available. 
Hence, we split the original dataset into training, validation, and test subsets (70\%/10\%/20\% split) for our experiments. 
We denote this dataset as \smalldota\footnote{To enable reproducibility and future comparison of results, the train-validation-test split of \smalldota is available at \url{https://github.com/ChungYi347/Interactive-Multi-Class-Tiny-Object-Detection}.}. 
Following ~\cite{Xia_2018_CVPR}, we generate a series of 1024 × 1024 pixel patches from the revised dataset with a stride of 512 pixels, and train our models to detect oriented bounding boxes (OBB).

\paragraph{\celldataset}

\celldataset is a private breast cancer histopathology dataset with bounding box annotations for 8 cell classes.
\celldataset consists of $768 \times 768$ size patches and has 3681 training, 250 validation, 823 test samples.
We show in \autoref{tab:datasets} that on average, the patches in \celldataset contain twice as many objects as in \smalldota.
Further information about \celldataset is provided in our supplementary materials.

\subsection{Evaluation Procedure}
\label{sec:evaluation-simulation}

The evaluation of an interactive annotation system is a challenging topic.
In the most ideal case, we could evaluate using a large number of human annotators over many data samples, but this is somewhat infeasible and certainly not reproducible.
We thus take inspiration from the evaluation procedure in \cite{xu2016deep} for interactive segmentation, which plots task performance against simulated user clicks\footnote{\cite{xu2016deep} also proposes a "mean number-of-clicks" metric, but this must be computed per-sample and cannot be done for the mAP metric.}.

We simulate up to 20 ``clicks'' per image sample. 
This simulation over the full test set is an \emph{evaluation session}. 
For each "click", we randomly sample an object from a given image's ground-truth (without replacement), taking the object's center position and class index as the simulated user input.
When all available ground-truth objects are provided as simulated user inputs (for an image sample), no further user inputs are provided (similar to the training-time sampling method in~\autoref{sec:training-simulation}). 
This results in a set of predicted bounding boxes for an increasing number of users' clicks.
At each step, mAP\footnote{We compute mAP with an IoU threshold of 0.5.} is computed over all test set predictions for the corresponding number of clicks, and a plot of mAP versus clicks can then be made.
We perform five independent \emph{evaluation sessions}, and show the means and standard deviations of each data point using error bars.

\begin{figure}[t]
  \centering
  \begin{subfigure}[b]{1.0\columnwidth}
    \centering
    \includegraphics[width=\columnwidth]{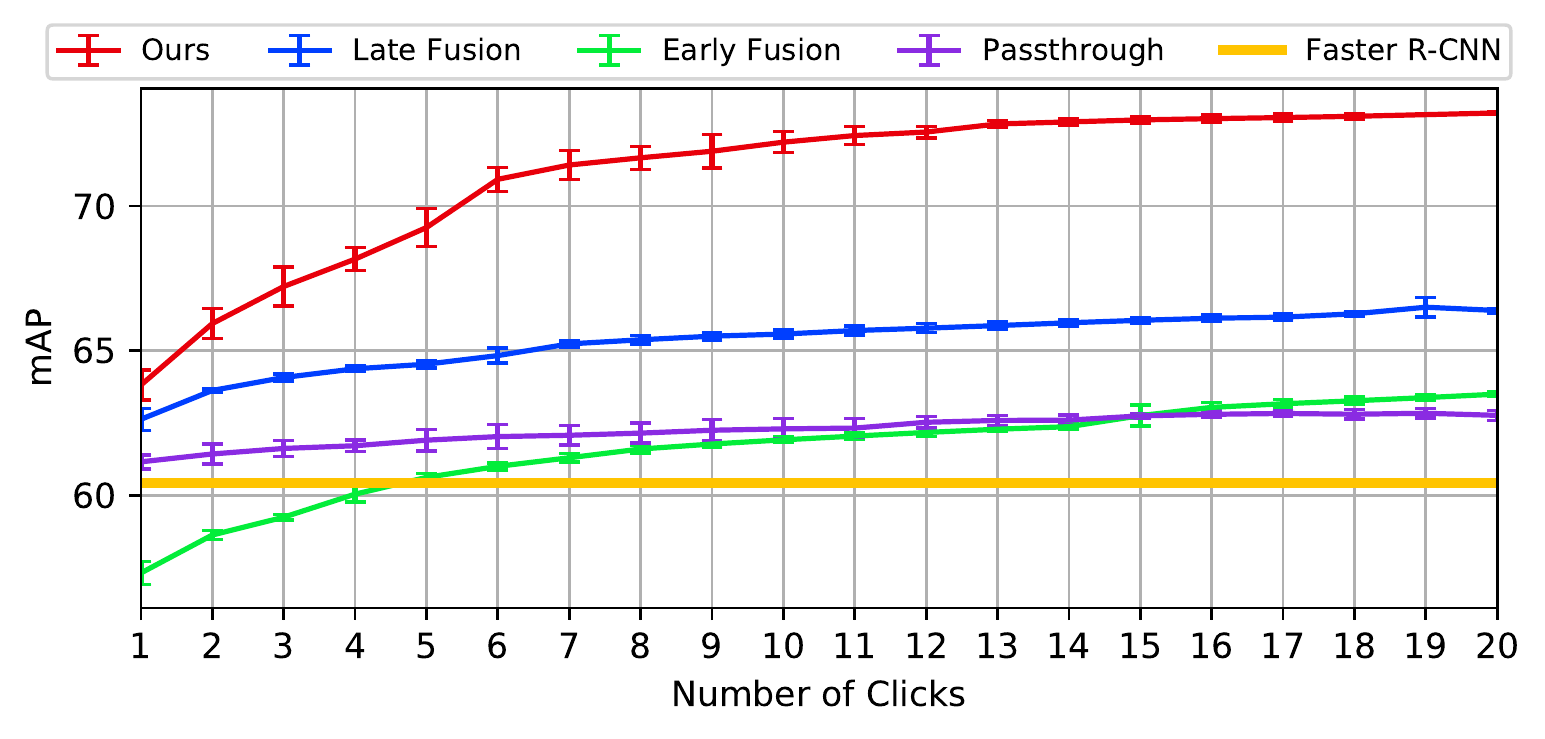}
    \vskip -1mm
    \caption{\smalldota}
    \label{baseline_fasterrcnn_dota}
  \end{subfigure}
  \vfill
  \begin{subfigure}[b]{1.0\columnwidth}
    \centering
    \includegraphics[width=\columnwidth]{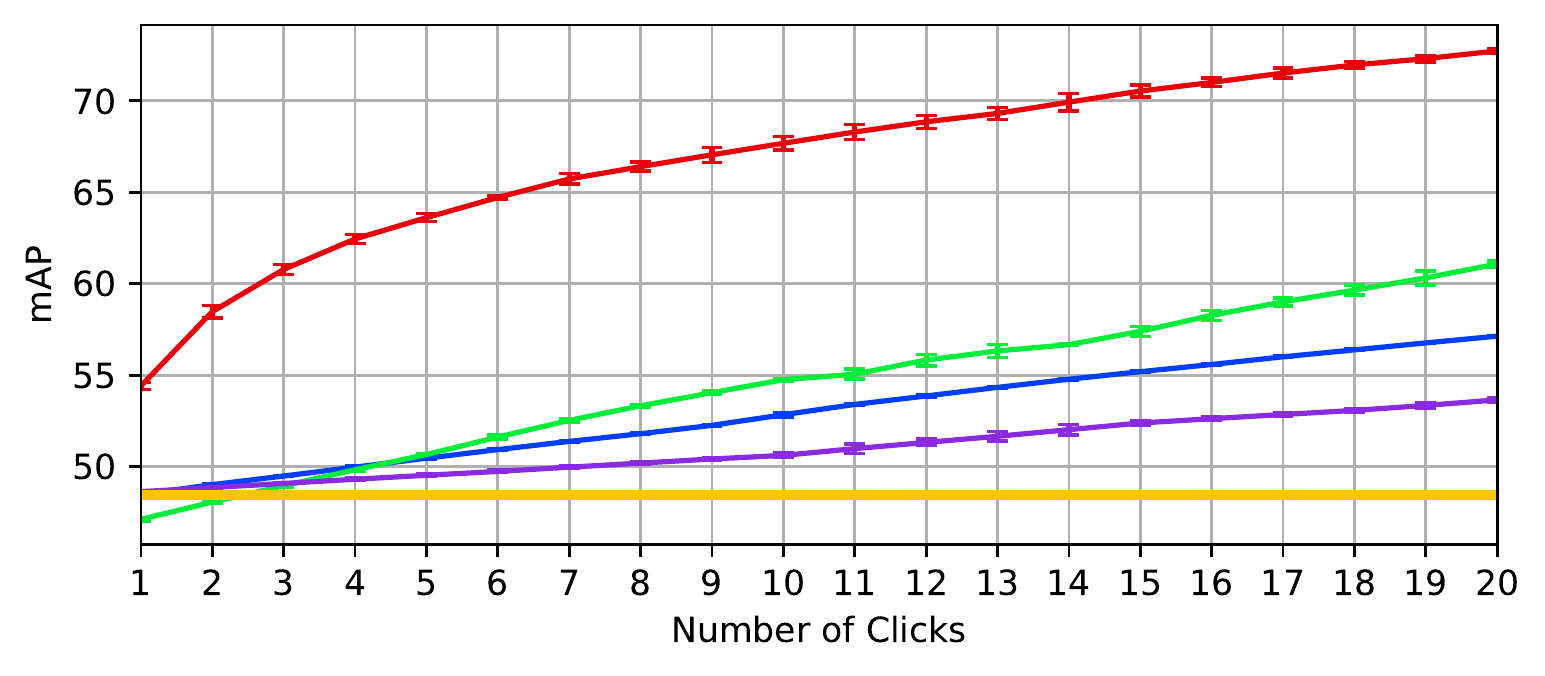}
    \vskip -1mm
    \caption{\celldataset}
    \label{baseline_fasterrcnn_cell}
  \end{subfigure}
  \vskip -3mm
  \caption{\netname (Faster R-CNN w/ R50-FPN) performance on \smalldota and \celldataset datasets compared to baseline methods.
    \label{fig:baseline_fasterrcnn}
  }
  \vspace{-0.5cm}
\end{figure}

\begin{figure}
    \centering
    \includegraphics[width=\columnwidth]{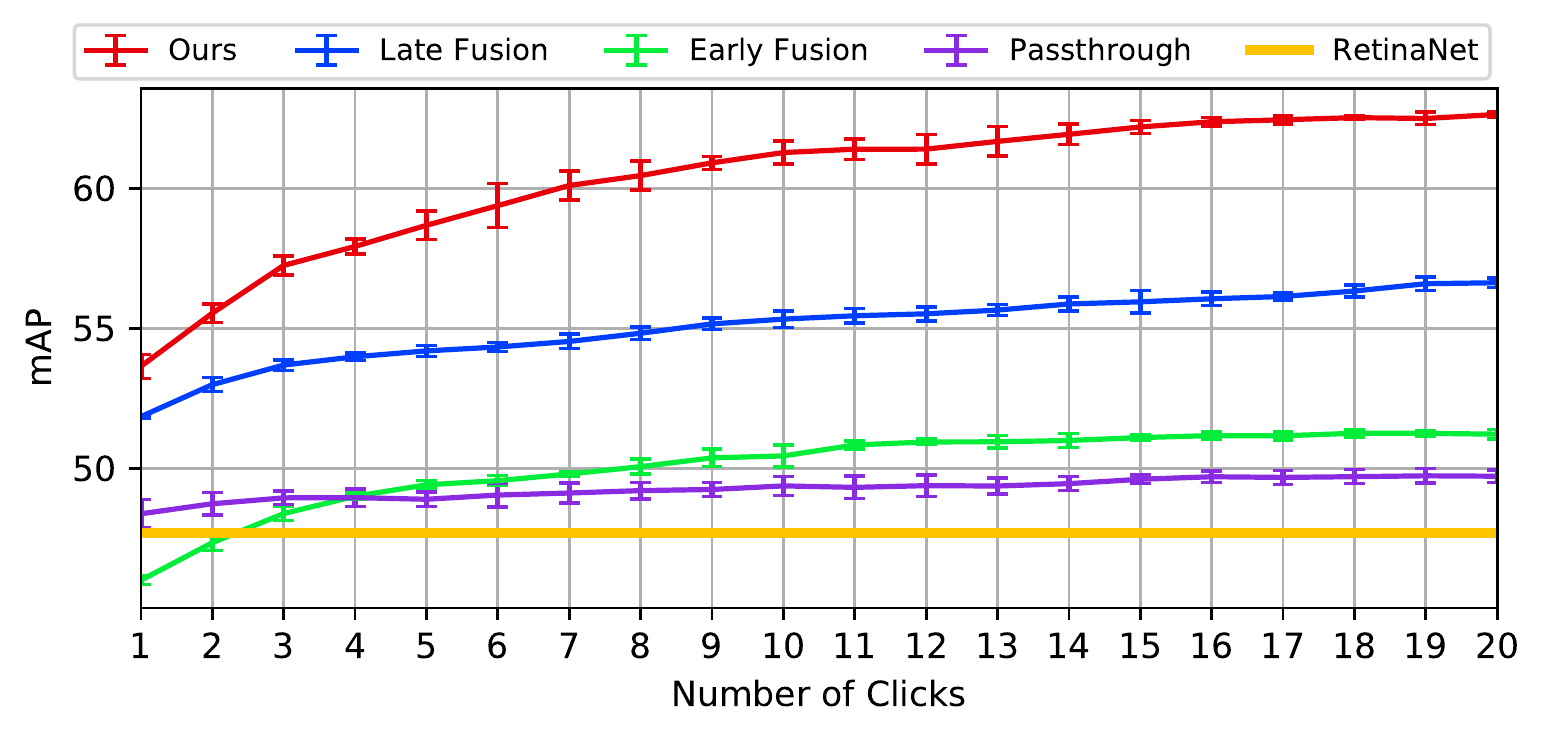}
    \vskip -3mm
    \caption{\netname (RetinaNet w/ R50) performance on \smalldota compared to baseline methods.
    \label{fig:baseline_retinanet_dota}
  }
  \vspace{-0.5cm}
\end{figure}

\subsection{Comparison to Baselines}

We compare our \netname approach against a few baseline methods in \autoref{fig:baseline_fasterrcnn}. 
For the compared methods, where applicable, we employ a Faster R-CNN architecture with a ResNet-50 (with feature-pyramid network) feature extractor~\cite{lin2017feature}. 
The lines labeled \emph{Faster R-CNN} in \autoref{fig:baseline_fasterrcnn} are the performance of the detector when simply trained on the labeled data without any interactive possibilities.
We refer to this as a \emph{baseline detector} in this section.
The compared methods are as follows:

\paragraph{Ours.} The full \netname approach including the \late and \corr modules as well as the \lossname loss.

\paragraph{Early Fusion.}
Early-fusion is a common method in interactive segmentation~\cite{xu2016deep,xu2017deep,jang2019interactive,sofiiuk2020f} and thus we implement it by concatenating the user-inputs heatmaps to the input image before passing through the feature extractor.
When drawing heatmaps, we use a larger $\sigma_\mathrm{heatmap}$ than other methods\footnote{For evaluating on \smalldota, we choose $\sigma_\mathrm{heatmap}=9$ for ``Early Fusion'', and $\sigma_\mathrm{heatmap}=1$ for ``Late Fusion'' and ``Ours''.}, as smaller Gaussians are less effective and their information can be lost in later layers. 

\paragraph{Late Fusion.}
Late-fusion is also commonly used in interactive segmentation~\cite{zhang2020interactive,agustsson2019interactive,rakelly2018few}, and is a competitive baseline method.
We implement this baseline by using our \late module but omitting the \corr module and \lossname loss.

\paragraph{Passthrough.} A naive yet effective baseline is one where the class value of user-inputs are simply applied to matching predicted bounding boxes from the baseline detector.

\paragraph{Results.}
We find that our proposed method out-performs all baselines consistently, quickly increasing in test set mAP with a few number of clicks, and reaching higher mAPs when the maximum number of clicks are provided.
The Early Fusion and Late Fusion baseline methods out-perform the naive passthrough method, but by smaller amounts compared to our approach.

\subsection{Application to a One-stage Detector}
Our approach can apply to both two-stage and one-stage detector architectures.
We show this by applying \netname to the RetinaNet architecture (with ResNet-50 backbone) and evaluating on \smalldota.
The one-stage results show similar tendencies as the two-stage case (\autoref{fig:baseline_fasterrcnn}), with the baselines showing modest improvements over the \emph{baseline detector}, and our method showing large improvements.
We thus show that our method can apply to both one-stage and two-stage architectures for object detection.

\begin{figure}
    \centering
    \includegraphics[width=1.0\columnwidth]{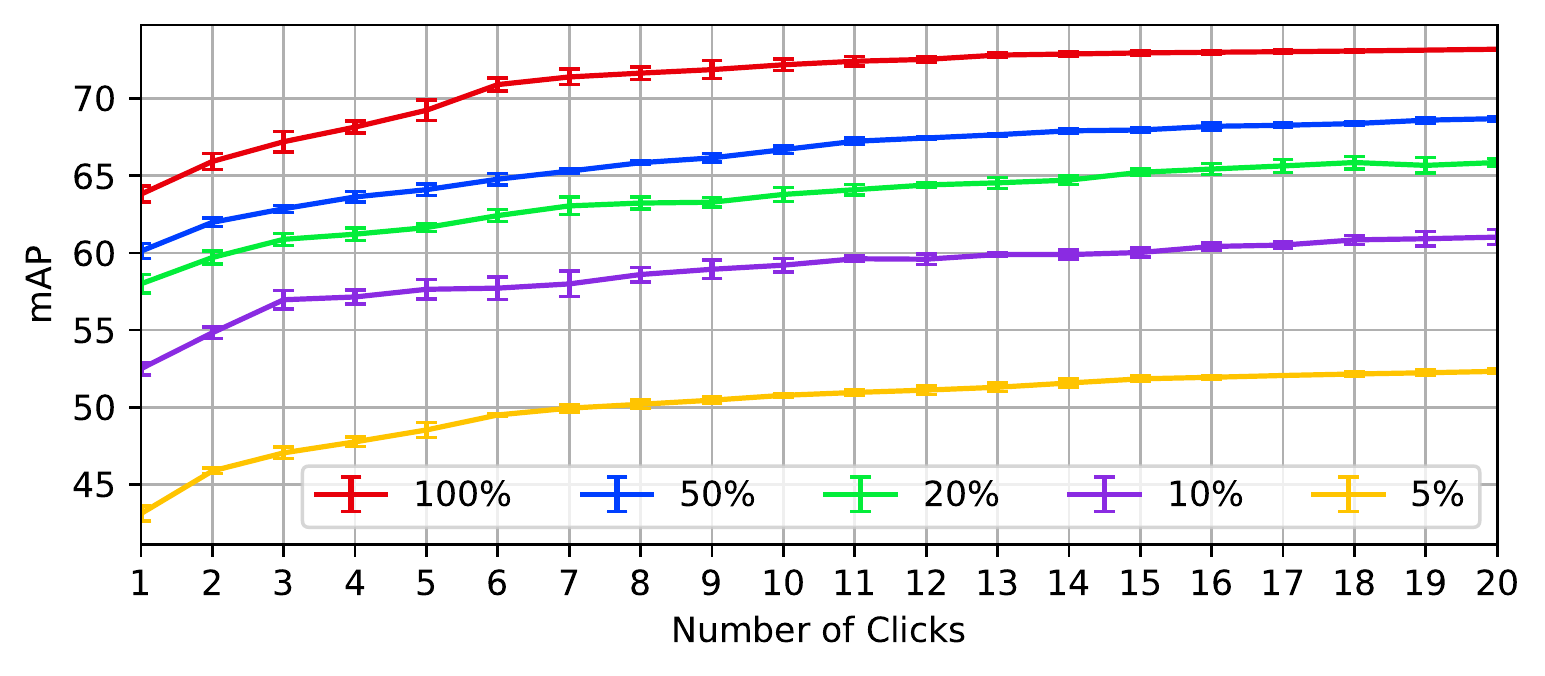}
    \vskip -2mm
    \caption{Decreasing the amount of training data (percentage of the full \smalldota training subset) still allows for \netname to increase annotation quality with increasing number of clicks.
    }
    \label{fig:dataset}
    \vspace{-0.5cm}
\end{figure}

\begin{figure*}
  \begin{subfigure}[b]{0.33\textwidth}
    \centering
    \includegraphics[width=\columnwidth]{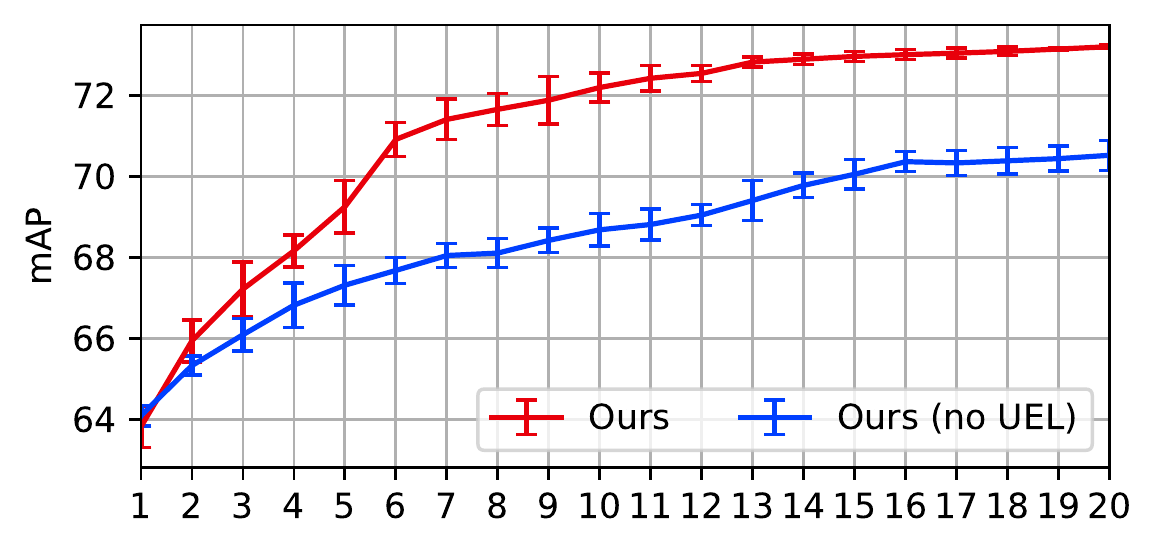}
    \caption{User-input Enforcing Loss (UEL)}
    \label{ablation_loss}
  \end{subfigure}
  \hfill
  \begin{subfigure}[b]{0.33\textwidth}
    \centering
    \includegraphics[width=\columnwidth]{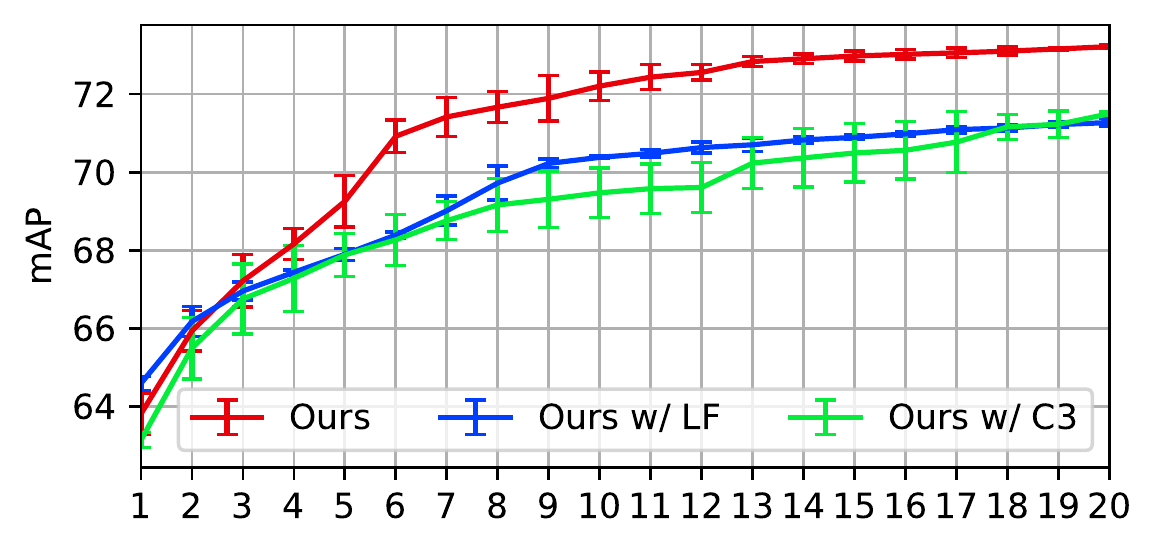}
    \caption{\late and \corr modules}
    \label{ablation_module}
  \end{subfigure}
  \hfill
  \begin{subfigure}[b]{0.33\textwidth}
    \centering
    \includegraphics[width=\columnwidth]{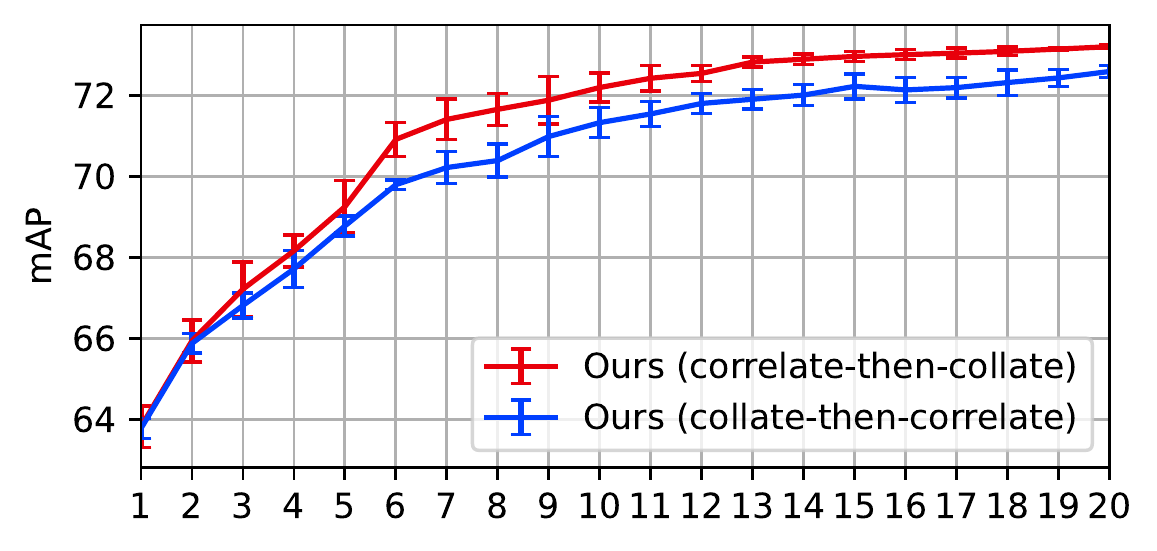}
    \caption{Correlation/Collation Order 
    }
    \label{ablation_corr_accum}
  \end{subfigure}
  \vskip -2mm
  \caption{\textbf{Ablation study} of \netname (Faster R-CNN w/ R50-FPN) on \smalldota. The graphs show the impacts of (a) User-input Enforcing Loss (UEL), (b) the combination of LF and C3 modules, and (c) the order of correlation and collation.
  }
  \vspace{-0.5cm}
\end{figure*}

\subsection{Varying Amount of Training Data}

In a real-world scenario, one may question whether our approach applies to cases with lower number of training samples.
We thus conduct an experiment by varying the amount of training data in \smalldota, using the Faster R-CNN architecture.
\autoref{fig:dataset} shows that our approach predicts bounding boxes with increasing mAP with increasing clicks, even with as little training data as 5\% (only 559 samples). 
In the real-world, then, a small set of fully-annotated data could be collected initially in order to train \netname.
This could then be used to assist annotators in labeling additional samples.
By repeating this process, even a large dataset could be annotated efficiently.

\subsection{Ablation Studies}\label{sec:ablation}
We conduct three ablation study on the \smalldota dataset to understand the impacts of our modules and loss.
The evaluated method is \netname based on the Faster R-CNN architecture with a ResNet-50 (with FPN) feature extractor.

\paragraph{User-input Enforcing Loss (UEL).}
\autoref{ablation_loss} shows that the addition of the user-input enforcing loss results in significantly better performance (especially for no. of clicks $>\,3$) compared to the case without \lossname.
It is clear that the UEL ensures that the consistency between user's inputs and the model predictions are improved.
This is demonstrated by the overall better performance both in the presence of few clicks as well as in the case of many clicks. 
Furthermore, the smaller error bars (std. dev. of mAP over 5 trials) at high no. of clicks, indicate that applying the UEL allows the model to better understand and incorporate user inputs overall, without being too sensitive to which user inputs specifically are provided.

\paragraph{\late Module and \corr Module.}

In \autoref{ablation_module}, we train models with \lossname and with either the \late module or \corr module, to compare the effect of the \late module against the effect of the \corr module.
We find that the proposed \late module and \corr module on their own show good performance overall.
However, it is when they are combined that a significant boost in performance is observed.
We hypothesize that this is because the \late module allows for the model to better understand the implication of user inputs, in the local areas around the input positions.
The \corr module on the other hand explicitly queries very far away objects for similarity.
In a manner of speaking, the \late module helps the model understand the local context in relation to user inputs, and the \corr module helps the model understand the global context. %
This holistic approach is beneficial, as is evident by the large boost in performance.

\paragraph{Class-wise Feature Correlation.}
Our \corr module performs feature correlation per user input, then merges the correlation maps by class (correlate-then-collate).
An alternative is to combine the user-input heatmaps class-wise first, then perform correlation (collate-then-correlate).
The latter approach assumes that all objects of a specified class are represented by similar ``template'' features.
In addition, it promises to be less sensitive to how well the user positions their input.
In contrast, the chosen approach (\corr) considers that objects from the same class can be represented by somewhat different features.
By performing correlations for each user-input, the \corr module embraces the within-class diversity of objects.
The results in \autoref{ablation_corr_accum} show that while both approaches work well, the correlate-then-collate method out-performs the collate-then-correlate alternative.

\section{User Study}
\label{sec:user_study}

To evaluate the efficacy of \netname in the real-world, we conduct a user study where annotators are asked to annotate OBB of objects on images taken from the \smalldota dataset.
We sample 40 images from the test set for this task, which contain 10 to 100 objects.
Our study is a within-subjects study, in which 10 participants perform their tasks with two conditions (in a random order).
The two task conditions are: (a) fully-manual annotation and (b) interactive annotation using \netname.
In the fully-manual case, annotators select an object class, then make 4 mouse clicks per object to draw a quadrilateral that is oriented based on the object's orientation.
In the interactive case, annotators are allowed to provide hints to \netname by (a) selecting an object class, and (b) clicking on an example object.
When the annotator is satisfied with \netname's predictions, they are then allowed to revise mislabeled objects and add missing objects via manual annotation.
Lastly, after completing each condition (manual or interactive), the annotators fill out the NASA-TLX questionnaire to assess their task load \cite{hart1988nasatlx}.

\begin{figure}[t]
  \begin{subfigure}[b]{0.32\columnwidth}
    \centering
    \includegraphics[width=\columnwidth]{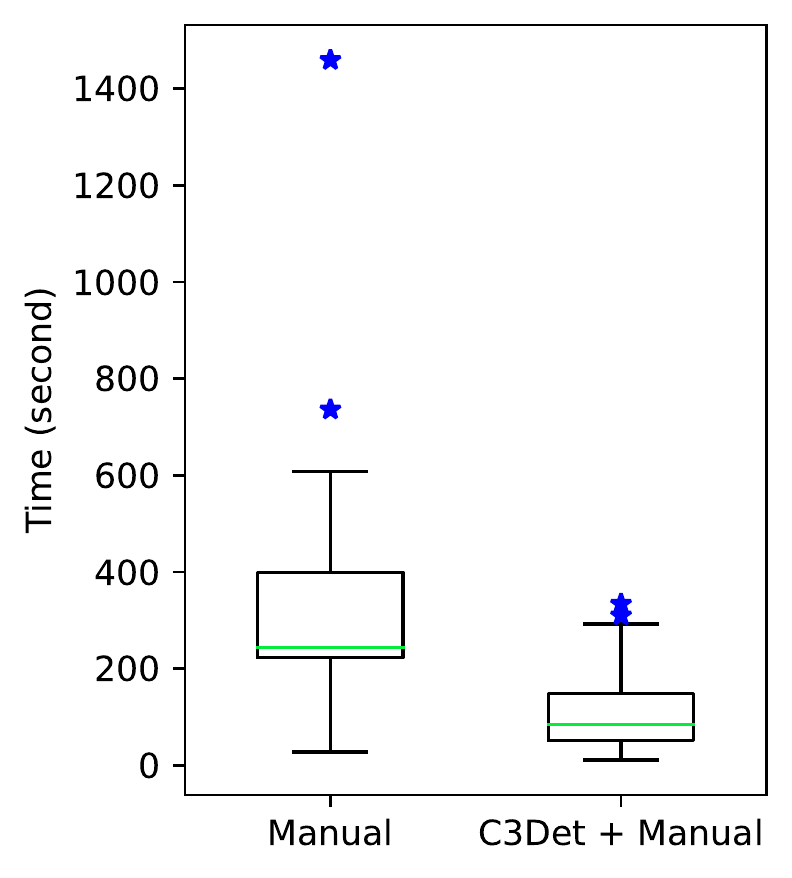}
    \caption{Time}
    \label{userstudy_time}
  \end{subfigure}
  \hfill  
  \begin{subfigure}[b]{0.32\columnwidth}
    \centering
    \includegraphics[width=\columnwidth]{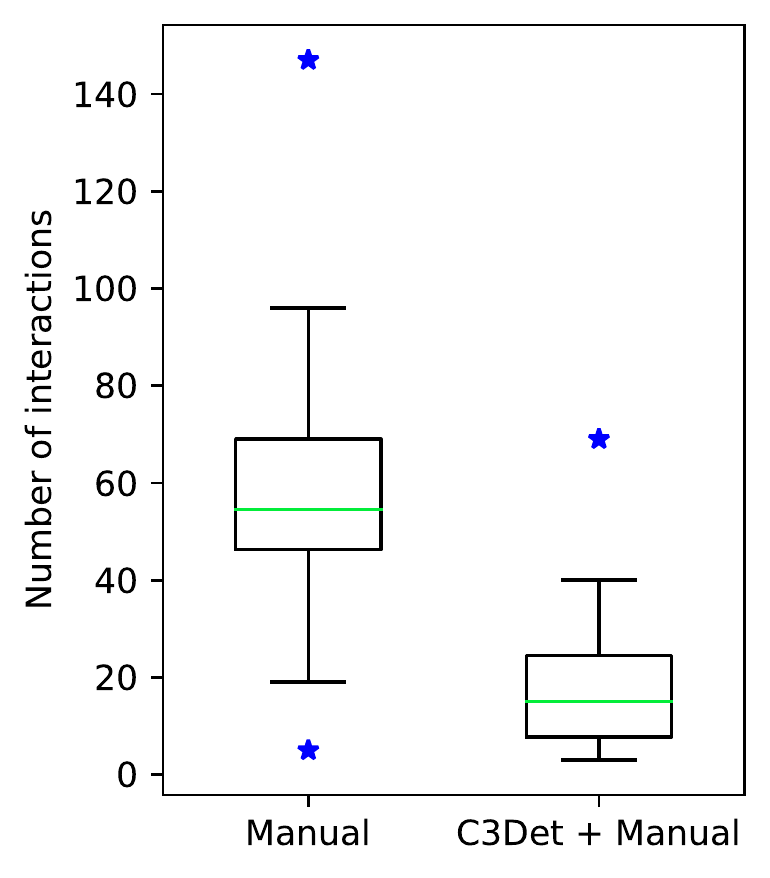}
    \caption{Interactions}
    \label{userstudy_interaction}
  \end{subfigure}
  \begin{subfigure}[b]{0.32\columnwidth}
    \centering
    \includegraphics[width=\columnwidth]{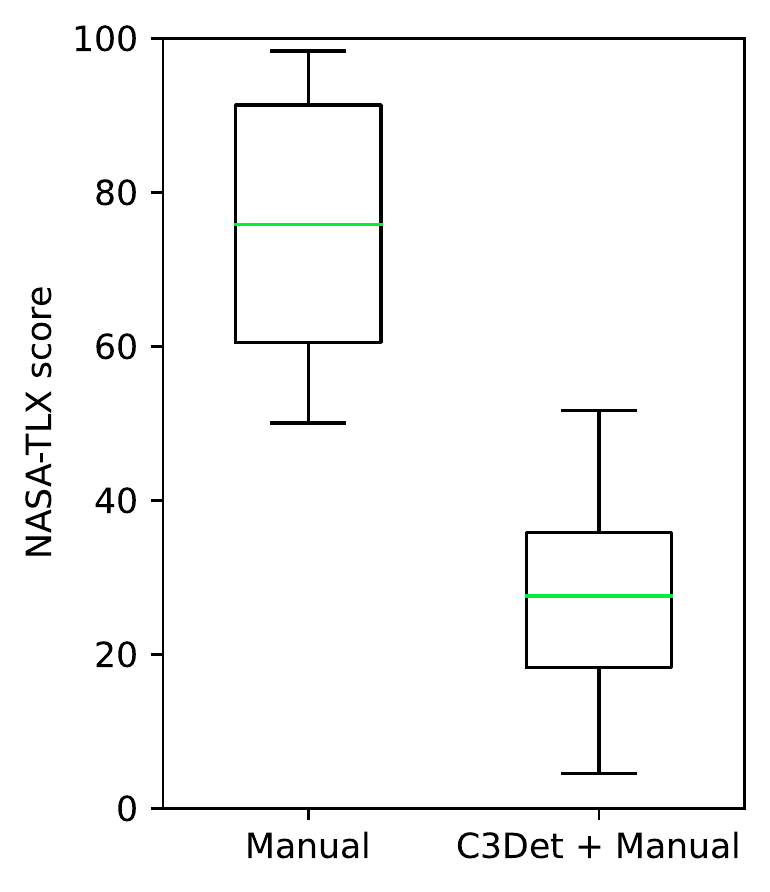}
    \caption{NASA-TLX}
    \label{userstudy_nasatlx}
  \end{subfigure}
  \vskip -2mm
  \caption{
  Box plots of per-sample (a) time taken and (b) number of interactions, and overall (c) task load assessed using NASA-TLX.
  }
  \vspace{-0.5cm}
\end{figure}

We analyze the annotation time and number of interaction for each condition, as shown in \autoref{userstudy_time} and \autoref{userstudy_interaction}.
In the interactive case, users are allowed to further modify the bounding boxes predicted by \netname, and so we call this the \emph{\netname + Manual} condition.
The fully-manual alternative is simply called the \emph{Manual} condition.
The average annotating time spent for \emph{\netname + Manual} (114.7s) is 2.85 times lower than the \emph{Manual} condition (327.73s). 
The number of interactions required for \emph{\netname + Manual} (17.93) is 3.25 times fewer than \emph{Manual} (58.33),
where interactions include the drawing and deleting of polygons, adding of user-inputs (for \netname), changing the selected class, and final result submission.
Please note that the time taken for \emph{\netname + Manual} includes model inference times, and shows that our method can work at interactive rates with annotators.

\autoref{userstudy_nasatlx} shows the NASA-TLX~\cite{hart1988nasatlx} score for each task session.
The median TLX score with the Manual approach is 75.83, and the score with \netname is 27.58 (lower is better). 
A paired Wilcoxon signed-rank test with a continuity correction is conducted and we find that the difference in task workload is statistically significant at a significance level of 0.01 ($z' = -2.703$, $p' = 0.0069$, $r' = 0.604$).
This means that annotating with \netname takes less time, interaction, and workload as measured by NASA-TLX.  

\begin{figure}
    \centering
    \includegraphics[width=0.9\columnwidth]{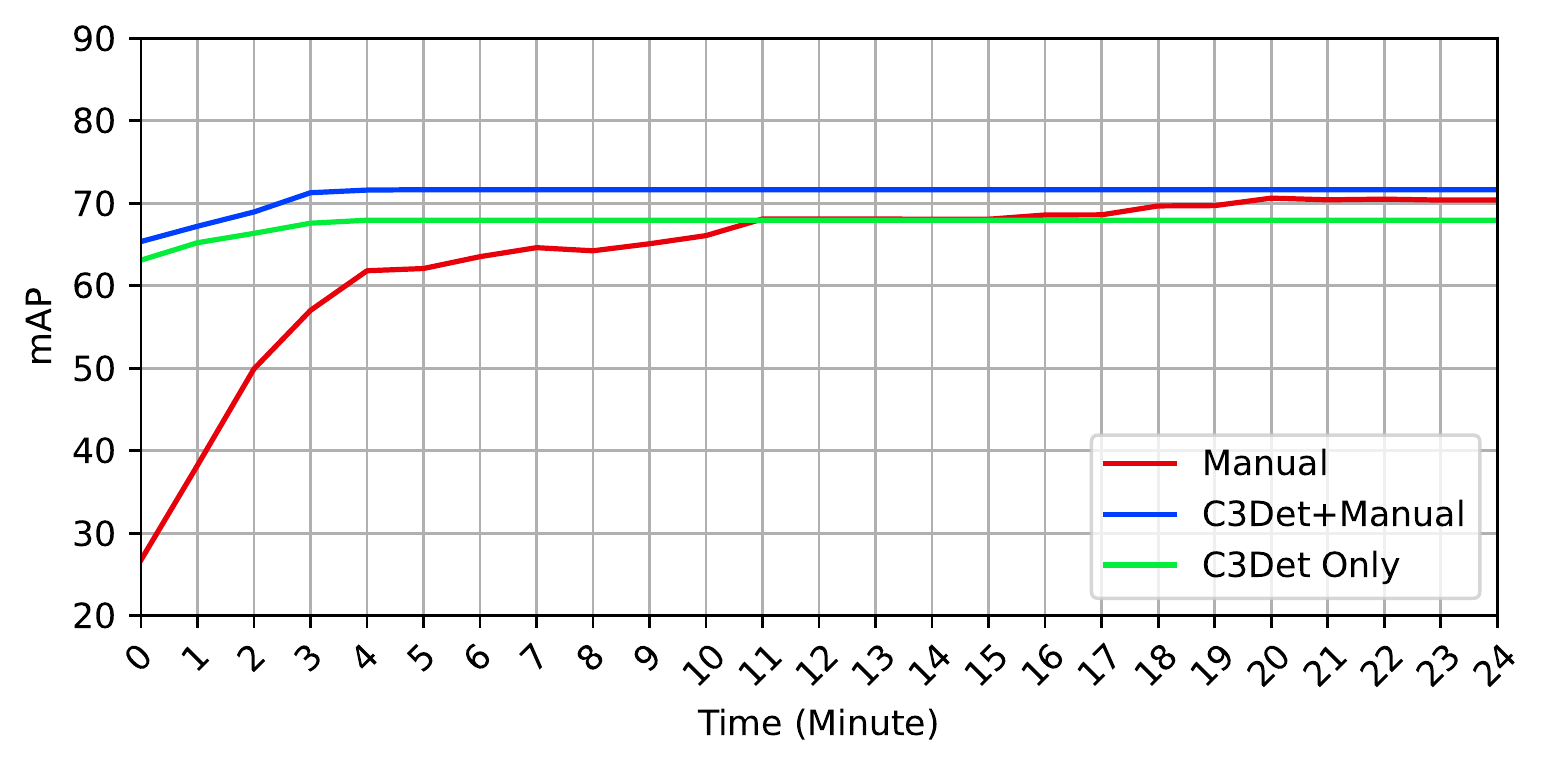}
    \vskip -3mm
    \caption{
    Annotation quality (mAP) versus annotation cost (time) for different annotation schemes. 
    Our \emph{\netname + Manual} reaches 67.9 mAP five times faster than \emph{Manual}.
    }
    \label{fig:userstudy_map}
    \vspace{-0.5cm}
\end{figure}

To evaluate the quality of the final annotations, we compute the mAP metric between the \smalldota ground-truth and the annotations acquired via our user study\footnote{Bounding box class-confidence scores are set to 1 to compute mAP.}.
We additionally introduce the \emph{\netname Only} condition, which is the annotation acquired only via interacting with \netname (without any  manual modifications by the user).
\autoref{fig:userstudy_map} shows the increase in mAP over time for the compared conditions.
When considering how long it takes to achieve 67.9 mAP, the \emph{Manual} condition takes 714.3s, while the \emph{\netname Only} and \emph{\netname + Manual} conditions take 294.2s and 144.2s respectively.
This shows that \emph{\netname Only} allows for faster annotation than \emph{Manual}, for comparable quality. 
Allowing further manual modifications in \emph{\netname + Manual} results in even better annotation quality, indicating that practitioners should consider allowing manual modifications even in interactive annotation systems.
Considering that our user study participants are not expert annotators for \smalldota imagery, we believe that our results also indicate that \emph{\netname + Manual} can be a effective system for novices.

\section{Conclusion}
We have shown that \netname is a compelling approach for interactive multi-class tiny object detection. 
\netname improves an annotation task which can otherwise be laborious and expensive.
Our novel architecture considers the local and global implication of given user inputs in a holistic manner.
A newly proposed training data simulation and an evaluation procedure for interactive multi-class tiny-object annotation defines a clear methodology for future work in this area.
Our experimental results and user study verify that our \netname outperforms existing approaches and can reduce cost while achieving high annotation quality. 
We hope that our approach alleviates concerns on annotation cost and workloads in the real-world (e.g., industry settings).

\paragraph{Limitations.}
In our work, we assume that annotators do not make mistakes when specifying object class. Therefore, \netname may not be robust to rogue annotators.
Separately, future work could propose an alternative to our point-based user inputs to target multiple objects with one interaction, further reducing annotation costs.

\paragraph{Potential Negative Impact.}
Our method improves annotation efficiency of imagery with many tiny objects. Unfortunately, surveillance imagery often contain many tiny objects and bad actors may benefit from our work.
On the other hand, research on climate change, farm crop monitoring, and cancer research are highly beneficial to human society, hopefully offsetting concerns related to surveillance.

{\small
\bibliographystyle{ieee_fullname}
\bibliography{arxiv}
}

\clearpage
\setcounter{section}{0}
\renewcommand{\thesection}{\Alph{section}}
\section*{\Large Appendix}

\noindent
\textbf{Overview.}
This supplementary material includes further information about the multi-class tiny-object datasets we evaluate on in the main paper, including class and patch statistics and images. We also provide additional implementation details of the approach described in the main paper. Finally, we share visualizations of annotations acquired from the user-study, and further discuss annotation quality in terms of mAP.

In addition to this document, we provide a demo video of the annotation tool that we implemented for our user study.
This annotation tool is capable of providing real-time feedback for interactive multi-class tiny-object detection, as demonstrated in the video.

\section{Datasets}

\paragraph{\smalldota}
We supplement the details regarding the \smalldota dataset by providing in \autoref{tab:dota_info}, information regarding the number of patches and the number of objects of each object class, for the subsets: train, validation, and test.
The conditions listed at \url{https://captain-whu.github.io/DOTA/dataset.html} stipulate that the annotations of the DOTA dataset (and in extension, the \smalldota dataset) are available for academic purposes only, with commercial use being prohibited.

\begin{table}[b]
\centering
\begin{tabular}{|l|l|l|l|}
\hline
                   & Train  & Val   & Test \\ \hline
Num. patches       & 11198  & 1692  & 2823 \\ 
Num. objs in total & 431056 & 72483 & 99766\\ \hline
Num. Plane         & 15161  & 2775  & 4236 \\ 
Num. Bridge        & 3908   & 671   & 673  \\ 
Num. SV            & 271252 & 41994 & 64612\\ 
Num. LV            & 46956  & 5412  & 8159 \\ 
Num. Ship          & 75835  & 18155 & 17723\\ 
Num. ST            & 12954  & 2534  & 3145 \\ 
Num. SP            & 3996   & 686   & 1096 \\ 
Num. HELO          & 994    & 256   & 122  \\ \hline
\end{tabular}
\caption{\textbf{Further statistics on \smalldota.} The abbreviated classes are: \emph{small-vehicle} (SV), \emph{large-vehicle} (LV), \emph{storage-tank} (ST), \emph{swimming-pool} (SP), and \emph{helicopter} (HELO).
}
\label{tab:dota_info}
\end{table}

\begin{table}[b]
\centering
\begin{tabular}{|l|l|l|l|}
\hline
                   & Train  & Val   & Test  \\ \hline
Num. patches       & 3423   & 234   & 821   \\ 
Num. slides        & 419    & 182   & 87    \\
Num. cell in total & 271434 & 19184 & 81745 \\ \hline
Num. LC            & 92973  & 7630  & 18708 \\ 
Num. Fi            & 43838  & 3128  & 11935 \\ 
Num. Ma            & 6414   & 365   & 1801  \\ 
Num. NG1           & 15496  & 1133  & 6066  \\ 
Num. NG2           & 61024  & 4008  & 29525 \\ 
Num. NG3           & 17222  & 867   & 5436  \\ 
Num. NTC           & 17168  & 1052  & 4938  \\ 
Num. EC            & 17299  & 1001  & 3336  \\ \hline
\end{tabular}
\caption{\textbf{Further statistics on \celldataset.} The annnotated classes are: 
lymphoplasma (LC),
fibroblast (Fi),
macrophage (Ma),
nuclear grade 1 (NG1),
nuclear grade 2 (NG2),
nuclear grade 3 (NG3),
necrotic tumor (NTC), and
endothelial cell (EC).
}
\label{tab:cell_info}
\end{table}

\begin{figure}[b]
  \centering
  \begin{tabular}{@{}c@{\enspace}c@{\enspace}l}
    \includegraphics[width=0.42\columnwidth]{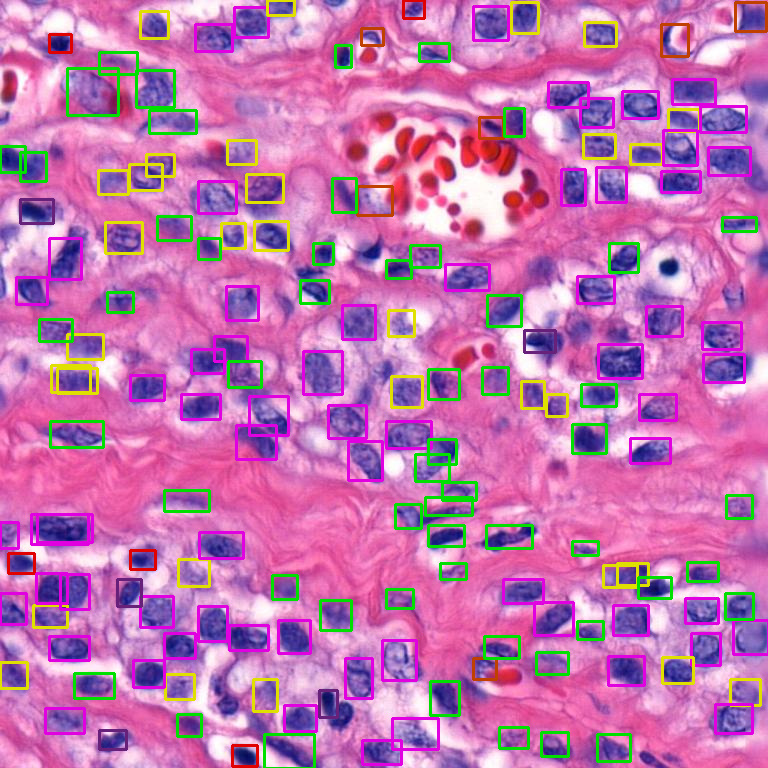} &
    \includegraphics[width=0.42\columnwidth]{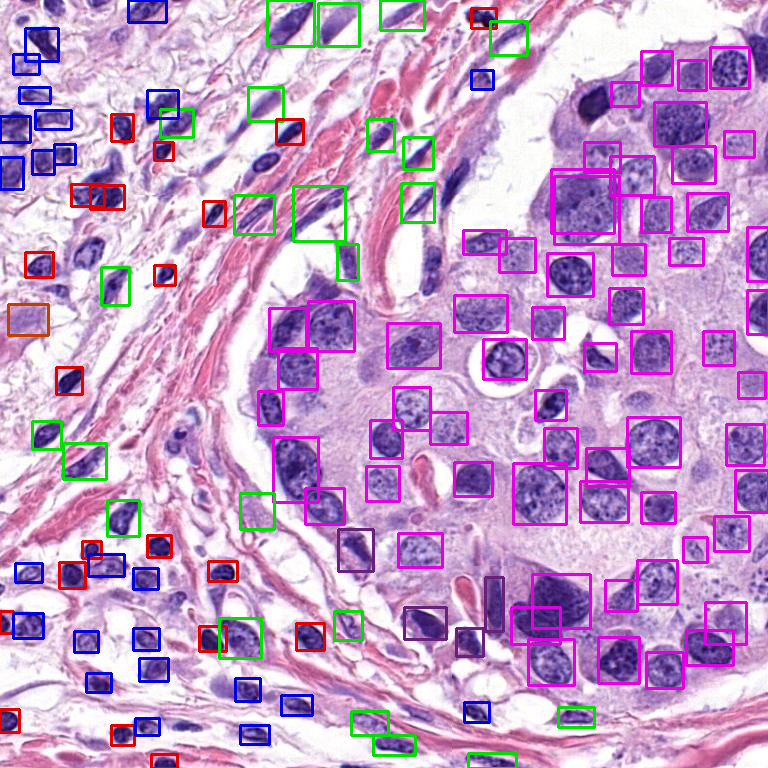} & 
    \includegraphics[width=0.13\columnwidth]{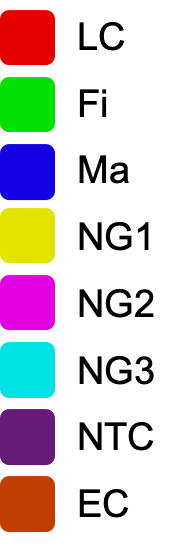} \\
    \includegraphics[width=0.42\columnwidth]{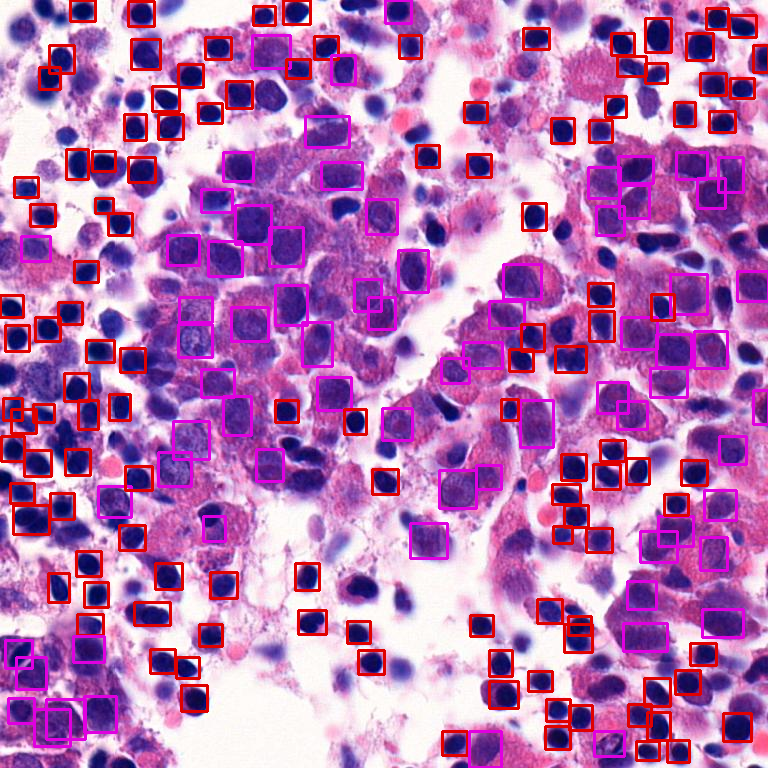} &
    \includegraphics[width=0.42\columnwidth]{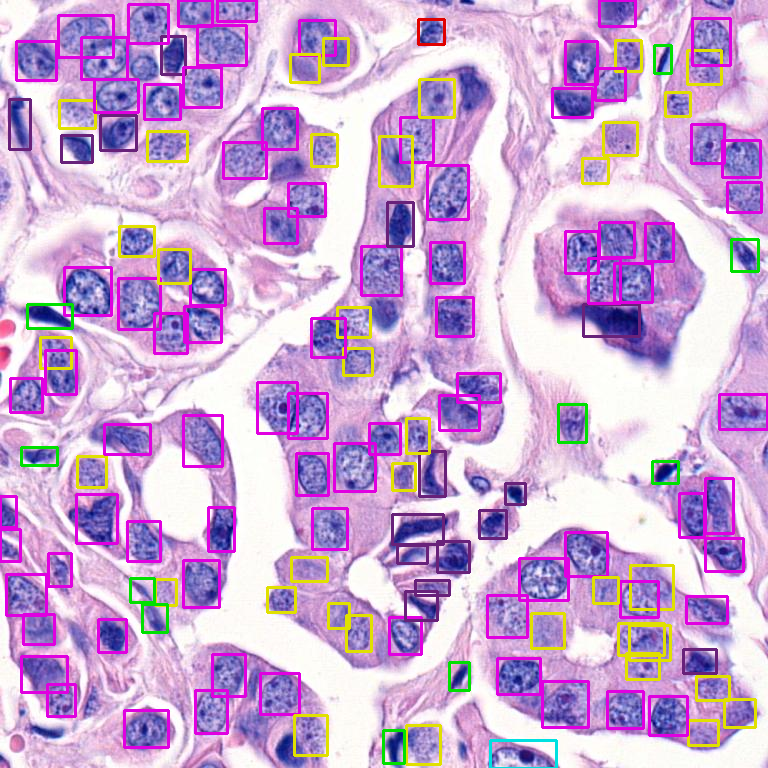} &
    \\
  \end{tabular}
  \caption{
  \textbf{Example images and annotations of \celldataset.}
  Each patch typically contains a large number of objects (cells), which may be challenging to distinguish.
  The annotation of such images require expert pathologists, who can benefit from an interactive annotation method that aids them in annotating many classes and objects from a few provided clicks.
  \netname promises to be such a method.
  \label{fig:cell_images}
  }
\end{figure}

\paragraph{\celldataset}
The \celldataset dataset is composed of patches taken from 688 whole-slide images of breast cancer biopsies\footnote{Due to the existing agreements regarding the whole-slide image data, we are unable to open-source this dataset, and it is therefore proprietary.}.
The patches were labeled by annotating breast cancer histopathology images for 8 cell classes.
The 8 cell classes annotated in our \celldataset include: lymphoplasma (LC), fibroblast (Fi), macrophage (Ma), nuclear grade 1 (NG1), nuclear grade 2 (NG2), nuclear grade 3 (NG3), necrotic tumor (NTC), and endothelial cell (EC). 
These classes are chosen by expert pathologists according to cells that are common in breast cancer histology. 
\autoref{tab:cell_info} states the number of patches, number of slides\footnote{We select fixed-size patches from several slides (whole-slide images), to use in training and validating models for cell detection.}, and the number of cells from each individual cell class.
We visualize some cell images along with their ground-truth bounding boxes in \autoref{fig:cell_images}.
Each bounding box represents a cell nucleus and the color denotes the annotated class of the cell.

\section{Model Implementation Details}

\paragraph{Feature Pyramid Network (FPN)}
Our approach and baseline models are trained with a ResNet-50 based Feature Pyramid Network (FPN \cite{lin2017feature}) as the CNN feature extractor (backbone) for Faster~R-CNN~\cite{ren2016faster} and RetinaNet~\cite{lin2017focal}, as mentioned in the \emph{``Experimental Results''} section.
In a nutshell, the FPN is a top-down architecture with lateral connections which uses multi-scale features better in detecting objects of various scales.
For each stage in the ResNet-based backbone, we forward the outputs through the FPN at the corresponding feature-map scale.
The number of output channels at each feature-map scale is 256 and the number of pyramid levels is 5.

\paragraph{Late Fusion (\late).}
Our CNN-based feature extractor (backbone) for processing user-input heatmaps is a ResNet-18.
Unlike in the case of the main task network (where we freeze the ImageNet-pretrained parameters up to the 1st ResNet stage), we do not freeze any layers of the \late module.
This is because user input heatmaps are very different in characteristic and number of channels compared to natural RGB images.
We also apply an FPN to the \late module, later concatenating features ($256$ features each) at matching feature-map scales from the main feature extractor.
The 512 feature maps ($2\times256=512$) are passed through a $1\times 1$ convolution layer to produce $256$ channels before using as an input to the RPN (in the case of Faster R-CNN), or the classification and box regression subnets (in the case of RetinaNet).
 
\paragraph{Training Configuration for \smalldota.}
We train Faster R-CNN and RetinaNet on \smalldota for 24 and 36 epochs, respectively.
We start from a learning rate of 0.01 (after a 500-step warmup) and scale it by 0.1x at 16 and 22 epochs (for Faster R-CNN), and 24 and 33 epochs (for RetinaNet), respectively. 

\paragraph{Training Configuration for \celldataset.}
On \celldataset, we train Faster R-CNN for 100 epochs. Training starts from a learning rate of 0.01 (after a 500-step warmup) and the is scaled by 0.1x at 30 and 60 epochs.

\begin{figure}[b]
    \centering
    \includegraphics[width=1.0\columnwidth]{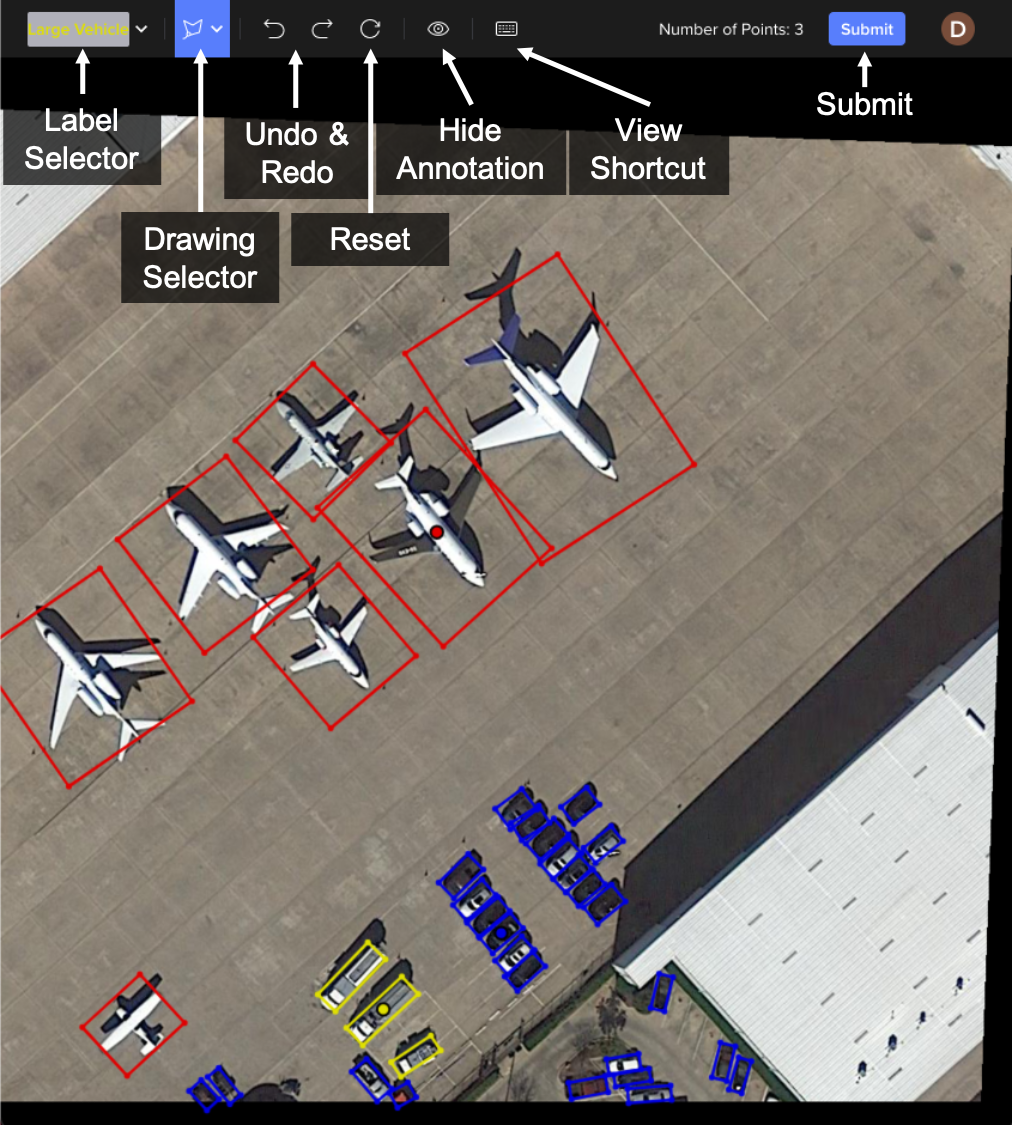}
    \caption{\textbf{User Study Frontend.} Our user-study GUI in semi-automatic annotation mode. Annotator inputs are shown as dots, while model predicted bounding boxes are drawn as quadrilaterals.
    }
    \label{fig:user_study_gui}
\end{figure}

\begin{figure*}[hbtp]
  \centering
  \vspace{9mm}
  \includegraphics[width=\textwidth]{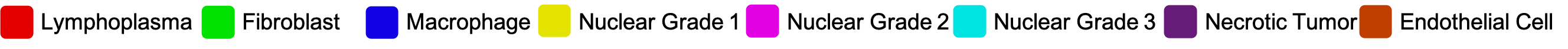} \\[2mm]
  \includegraphics[width=\textwidth]{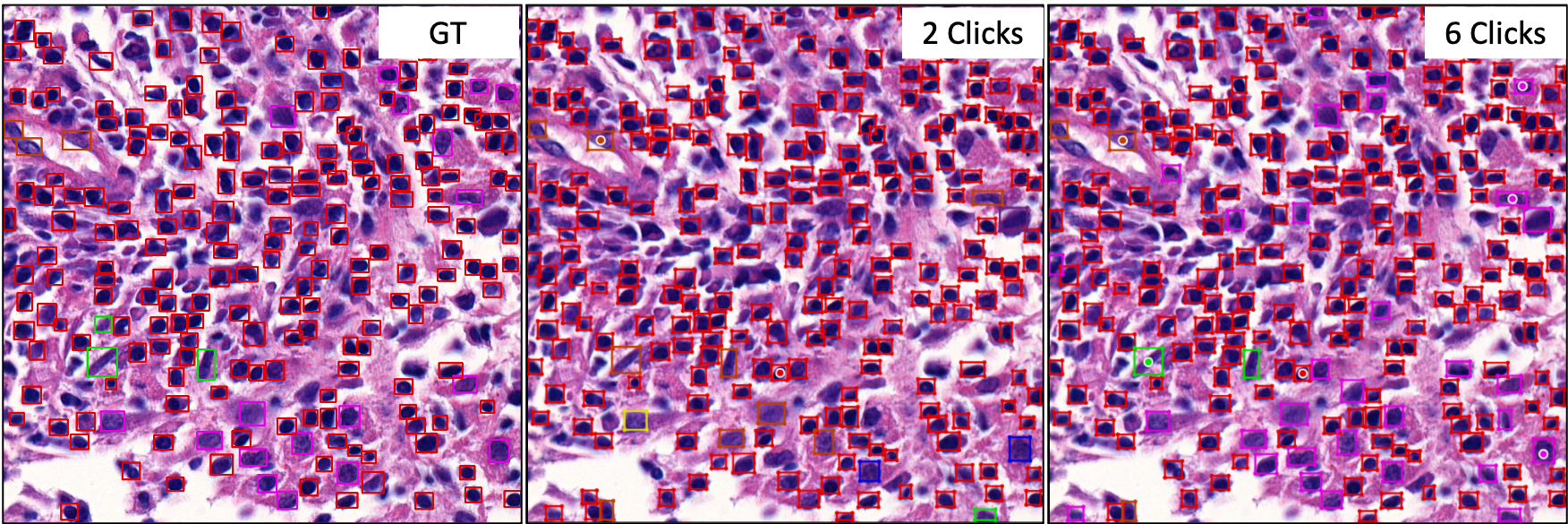} \\[2mm]
  \includegraphics[width=\textwidth]{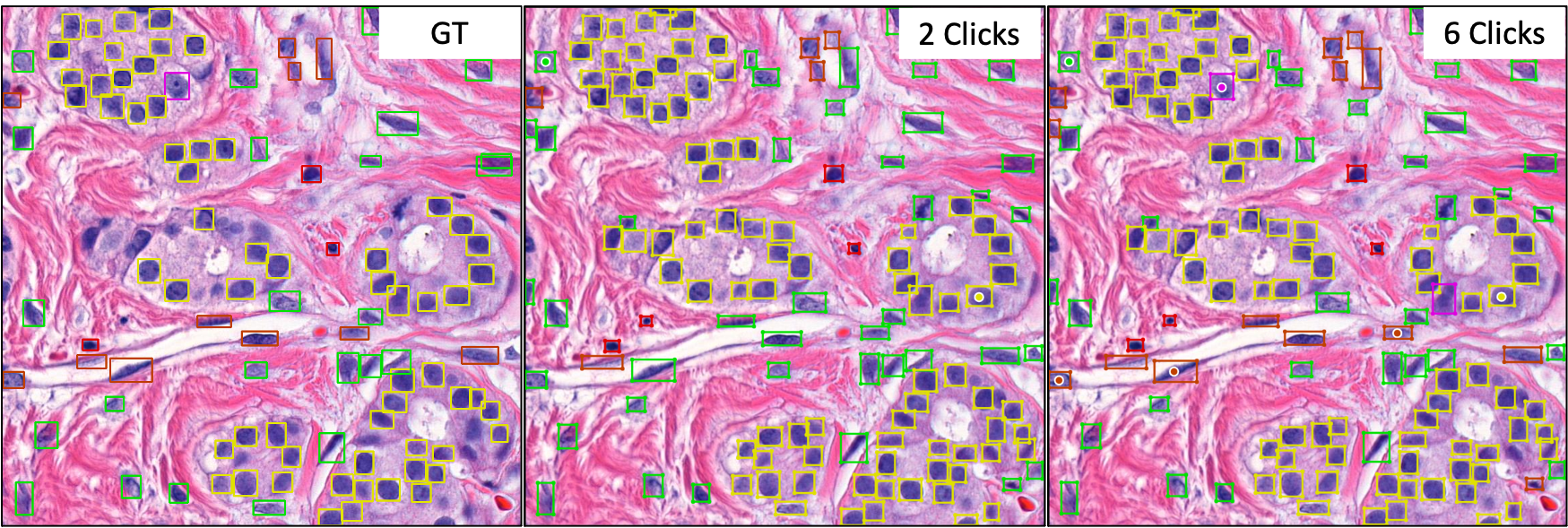} \\[2mm]
  \includegraphics[width=\textwidth]{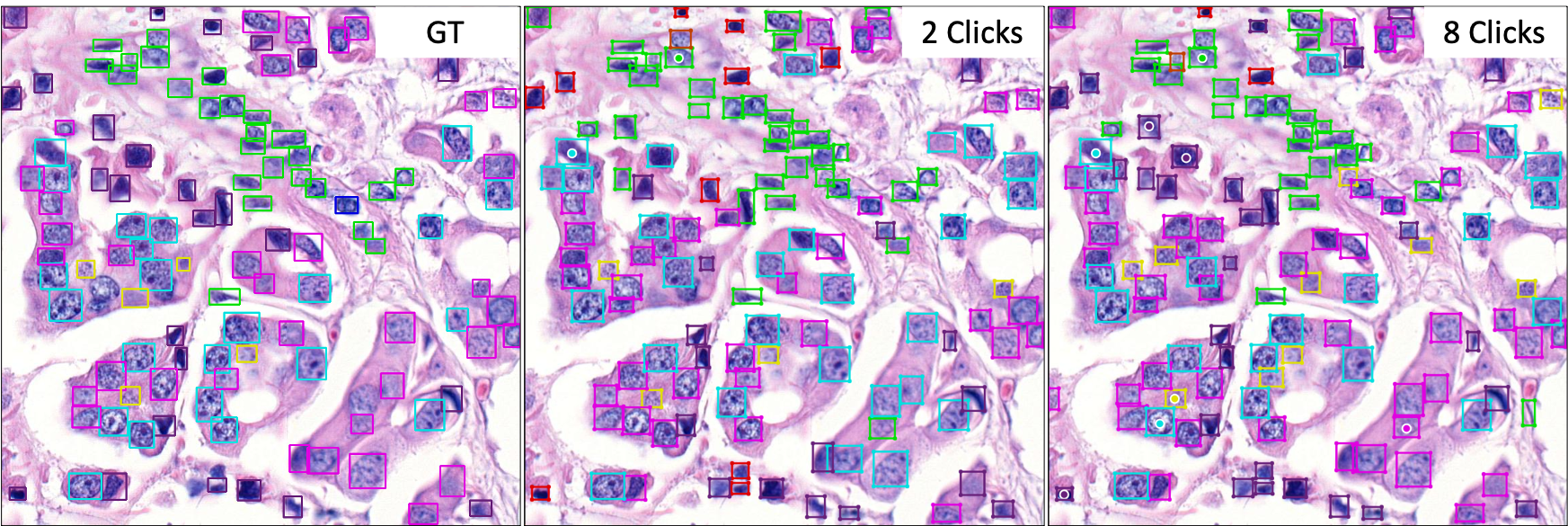} \\[2mm]
  \caption{
  \textbf{Visualizing Interactive Detection on \celldataset.}
  Example images of \celldataset with several prediction results by giving user inputs and ground-truth.
  The boxes and dots represents final annotated objects and user inputs from real annotators.
  \label{fig:cell_interactive}
  }
  \vspace{9mm}
\end{figure*}

\section{Interactive detection on \celldataset}
The \celldataset dataset is a particularly challenging dataset to annotate, requiring expertise and a large number of annotations per patch.
An effective interactive annotation method can easily reduce the necessary mouse clicks per patch from hundreds (no. of objects times 4) to just a few clicks.
We demonstrate that \netname can achieve this in \autoref{fig:cell_interactive}.

\autoref{fig:cell_interactive} illustrates 3 example cases from the \celldataset dataset, each with an increasing number of user inputs (described as clicks).
Even with a few given user clicks, object classes that were not explicitly provided in the user input are detected.
With more user inputs (up to 6 or 8), we find that the ground-truth can almost be reproduced.

In addition, we find that objects from the same class that are far away from the click position are detected.
As argued in the main paper, we believe that \netname considers local (nearby objects) and global context (far away objects) in relation to the input image and user inputs well, and address the multi-class tiny-object detection setting appropriately.

\begin{figure}[t]
  \centering
  \includegraphics[width=1.0\linewidth]{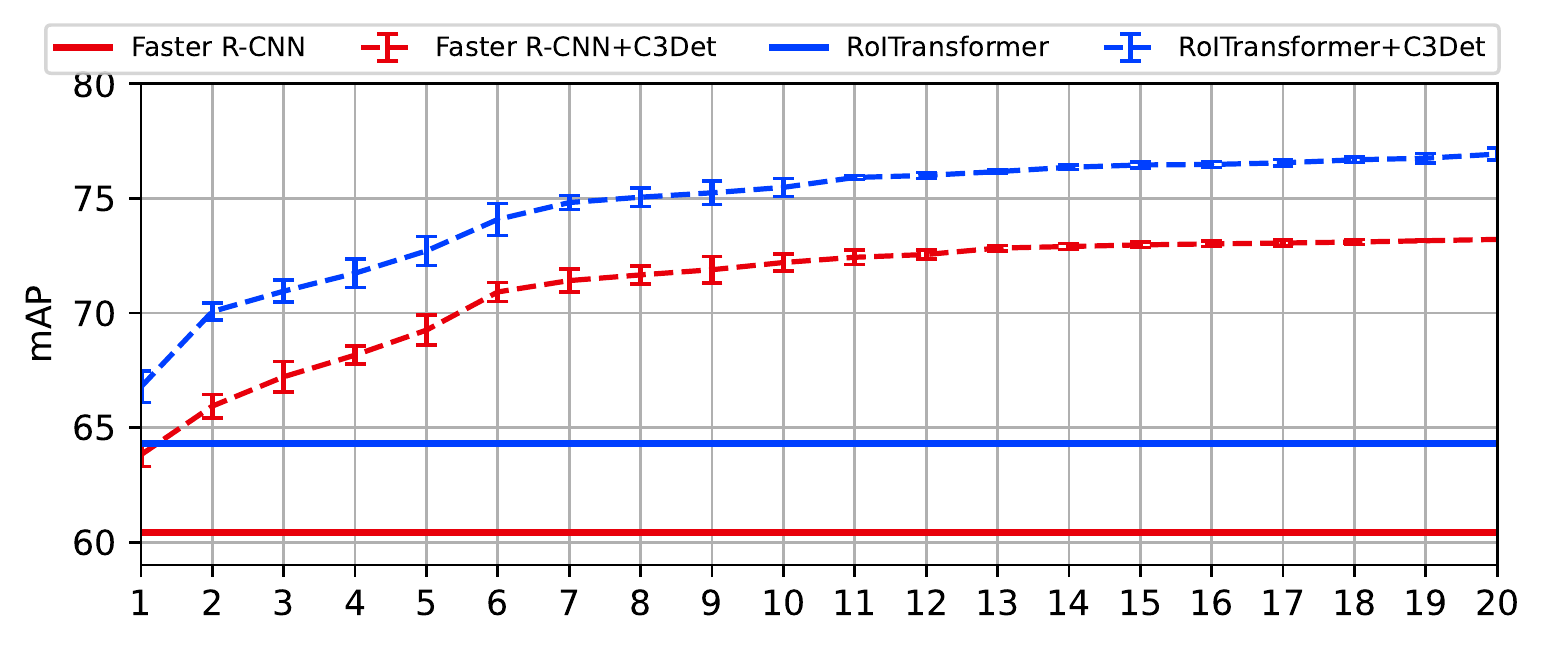}
  \includegraphics[width=1.0\linewidth]{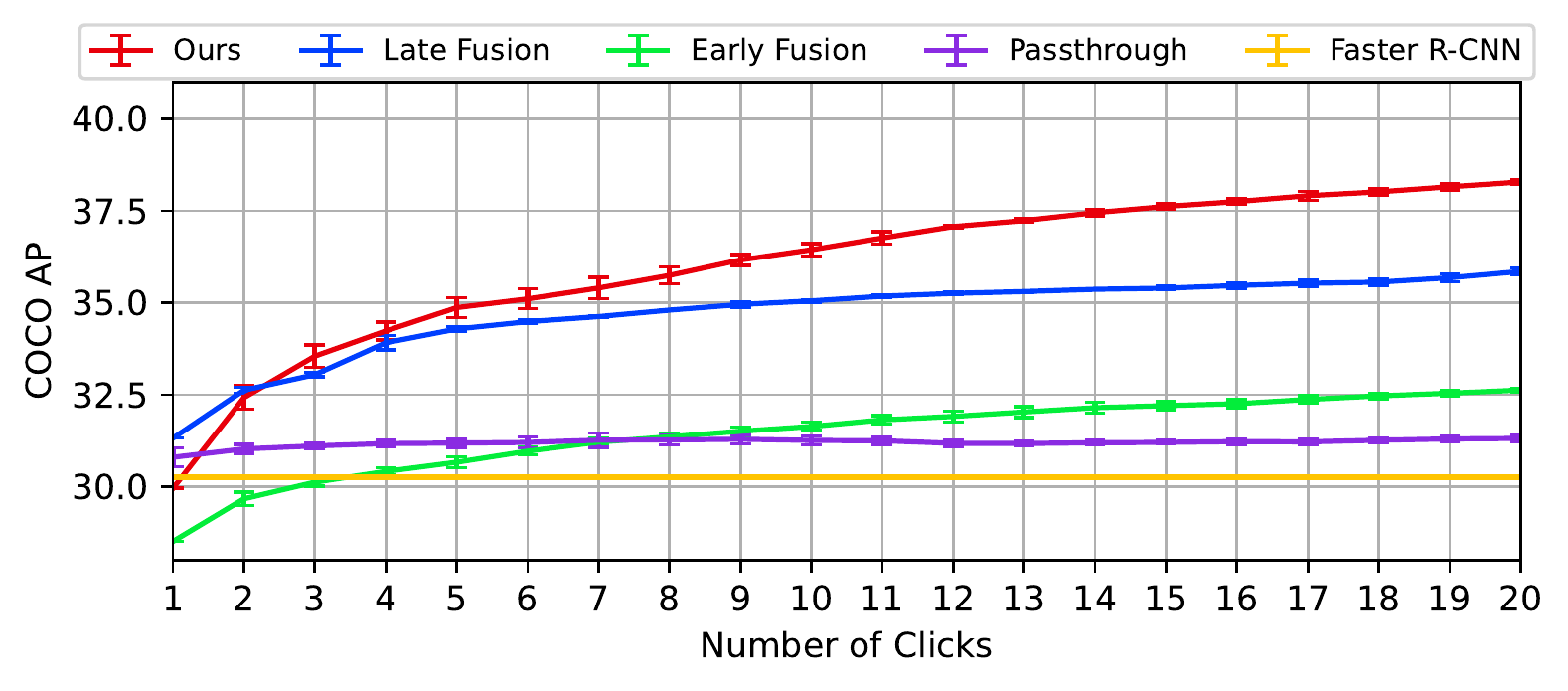}
  \caption{Performance on Tiny-DOTA. 
  \textbf{Top:} Both Faster R-CNN and RoI Transformer~\cite{ding2019learning} (solid lines) are improved by adding \netname (dashed lines). 
  \textbf{Bottom:} Our method against Faster R-CNN baselines measured with COCO AP@[.50:.05:.95]. 
  }
  \label{fig:onecol}
  \vspace*{-0.6cm}
\end{figure}

\section{Additional Experimental Results}
\paragraph{Comparison to another Tiny Object Baseline (RoI Transformer).}
Faster R-CNN and RetinaNet are standard baselines for detecting oriented bounding boxes on the DOTA dataset and are adopted in many recent papers. 
Therefore the results reported in the main paper suggest that our \netname can apply to variations of these standard networks.
Here, we demonstrate that this is indeed the case by selecting RoI Transformer~\cite{ding2019learning}, a recent detection method designed to perform well on DOTA.
Adding \netname to the purpose-built RoI Transformer still results in a significant and consistent increase in performance on \smalldota (see \autoref{fig:onecol}, top).
This further demonstrates the value of our proposed approach.

\paragraph{Performance on COCO AP metric.}
An IoU threshold of 0.5 is the standard procedure for evaluating on the DOTA dataset and thus we follow it in our experiments.
We believe, however, that it is also meaningful to compute the COCO AP metric, which takes into account objects of varying scales and therefore show this in ~\autoref{fig:onecol} (bottom). 
In comparison to Fig. 5a in main paper, we find that similar trends can be seen, though the numeric values are decreased due to the difficulty of the tiny objects task at a high IoU threshold.

\section{User Study}
We present a few more results and visualizations from our User Study.
In our user study, we asked participants to annotate images from the \smalldota dataset, using a fully-manual (\textit{Manual} condition) or semi-automatic (\textit{\netname + Manual} condition) approach.

At the beginning of each user study session, we asked for consent from the participant for their participation as well as the storing of their annotation and mouse clicks.
As no personally identifiable information was collected in our user study, we do not require approval from an institutional review board (IRB).

\paragraph{Implementation Details.}
We describe here the server specification and library used for implementing and serving the annotation tool used in our user study.

Our user-study GUI (see \autoref{fig:user_study_gui}) is implemented using several libraries such as FastAPI \footnote{https://fastapi.tiangolo.com/} for the back-end and React \footnote{https://reactjs.org/}, Redux-Saga \footnote{https://redux-saga.js.org}, and TypeScript \footnote{https://www.typescriptlang.org} for the front-end.
The images (to-be-annotated) are drawn on an HTML5 canvas using OpenSeadragon \footnote{https://openseadragon.github.io/} for convenient zooming and padding.
Likewise, user-inputs (points) and annotations (bounding boxes) are drawn using basic canvas methods.
Model inference via PyTorch takes only a few seconds (on a Titan X (Pascal) GPU) and we further show this real-time capability in a supplementary demo video.

\paragraph{Comparison of annotated images on \smalldota.}
\autoref{fig:dota_examples} shows example images with ground-truth annotations as well as annotations acquired by our user study conditions: fully-manual (\textit{Manual} condition) or semi-automatic (\textit{\netname + Manual} condition).
Compared to the ground-truth, the \textit{Manual} condition and \textit{\netname + Manual} condition achieve good quality, with small objects being annotated well.
However, there are few difference between them due to confusing object (in terms of object class), misconception of class definition and overly small objects. 
\autoref{fig:dota_examples_a} and \autoref{fig:dota_examples_b} are compelling example of frustrating objects (broken \emph{plane} and \emph{helicopter}) and misunderstanding of classes (\emph{small-vehicle} confused as \emph{large-vehicle}).
On the other hand, \autoref{fig:dota_examples_c} has many \emph{small-vehicle} and \emph{ship} objects in the bottom-right and top-left part of the image, respectively.
Although the original DOTA dataset does not have annotations for those very tiny objects (does not exist in the ground-truth), our annotator labeled these as small-vehicle objects (\netname + Manual).
In some manner, our semi-automatic approach may be reducing required effort, and allowing for more rich annotations to be produced.

\begin{figure}[]
    \centering
    \begin{subfigure}{1.0\columnwidth}
        \includegraphics[width=\columnwidth]{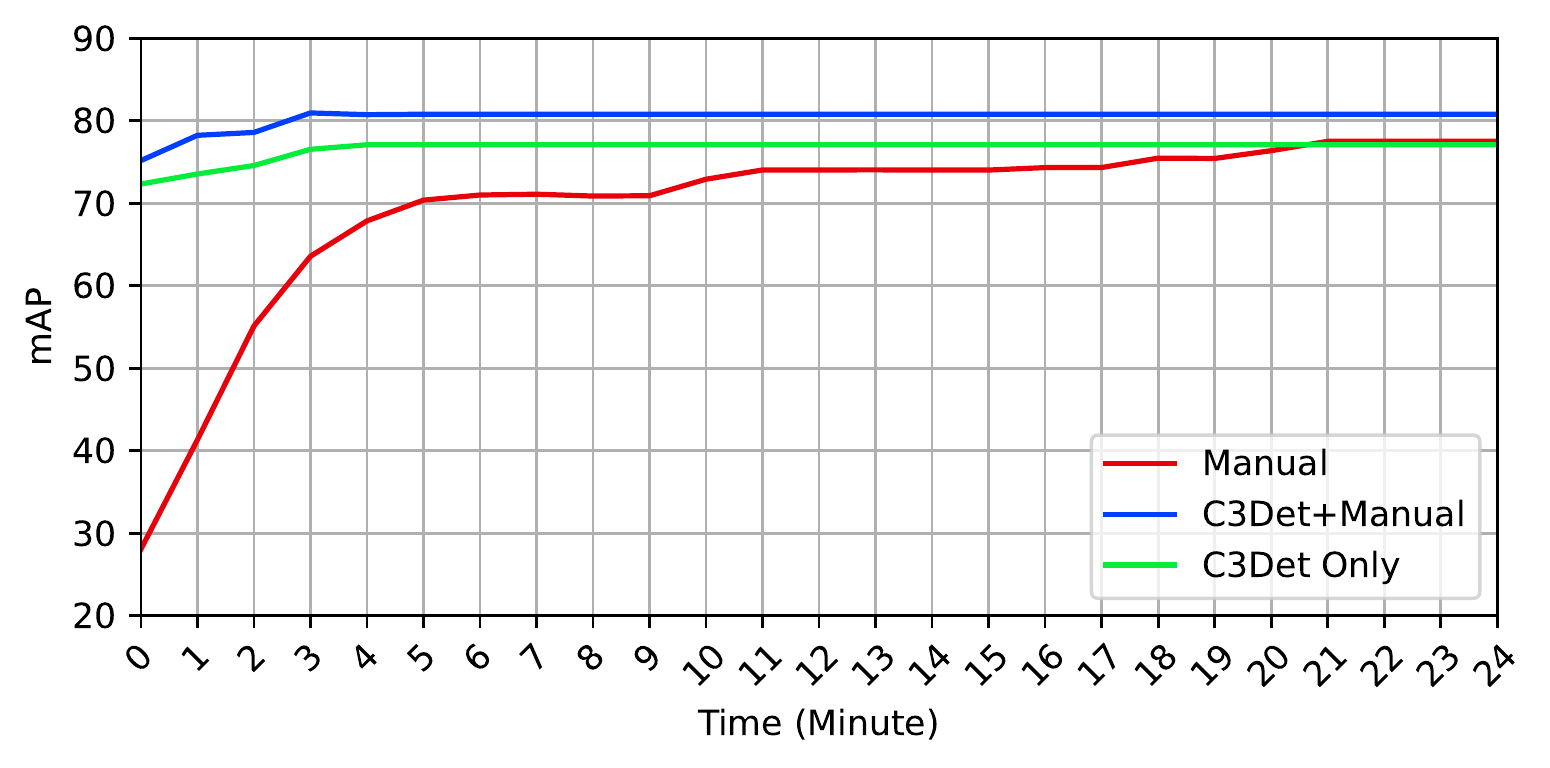}
        \caption{IoU threshold for mAP = 0.1}
        \label{fig:map01}
    \end{subfigure}
    \hfill
    \begin{subfigure}{1.0\columnwidth}
        \includegraphics[width=\columnwidth]{figures/map.pdf}
        \caption{IoU threshold for mAP = 0.5}
        \label{fig:map05}
    \end{subfigure}
    \caption{\textbf{Annotation quality (mAP) versus annotation cost (time).} for different annotation schemes in the user study. 
    Lowering the IoU threshold (\textbf{top}) for calculating mAP allows more loosely drawn bounding boxes to become valid annotations as well, compared to the results shown in the main paper (\textbf{bottom}).
    In fact, the \textit{\netname + Manual} condition produces better annotations overall than the \textit{Manual} condition.
    We believe that this may partly be due to the novice-level expertise of our annotators.
    \label{fig:user_study_mAP}
}
\end{figure}

\paragraph{Further Evaluation of Annotation quality (mAP).}
A typical assessment of the accuracy of bounding boxes is via the calculation of the mAP metric, with true-positives being assessed based on an intersection-over-union (IoU) threshold of 0.5 between a prediction and corresponding ground-truth box.
We therefore evaluated the annotation quality yielded by the different conditions in our user study using an IoU threshold of 0.5.

However, our user study participants are novices, and with the added complexity of drawing (often) very small bounding boxes using a computer mouse, we find that the acquired annotations were not always sufficiently covering the tiny objects (in particular, classes such as \emph{small-vehicle} suffered from this issue).
We therefore evaluate the mAP of acquired annotations with an IoU threshold of 0.1 and report it in \autoref{fig:user_study_mAP} (top).

In line with our observation, we find that the mAP increases for all annotation conditions.
In particular, our \emph{\netname + Manual} is able to yield better annotations overall than the \emph{Manual} or \emph{\netname Only} conditions.
This eludes to two possibilities:
\begin{inparaenum}[(a)]
\item \netname can guide novice annotators to produce better quality annotations, and
\item allowing manual edits on top of \netname outputs allows for an even higher final annotation quality.
\end{inparaenum}

\begin{figure*}[t]
  \centering
  \includegraphics[width=\textwidth]{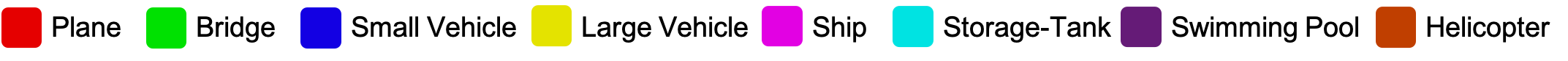} \\[2mm]
  \begin{subfigure}{\textwidth}
      \includegraphics[width=\textwidth]{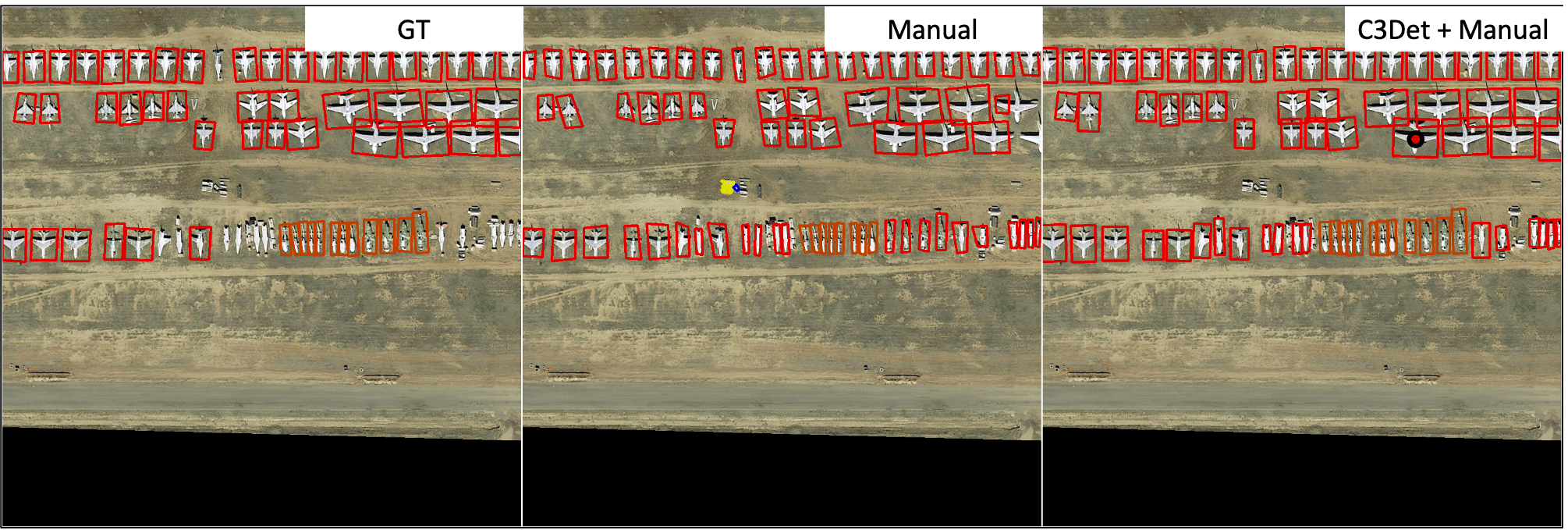}
      \caption{In some cases, the ground-truth (GT) omitted some objects, which our user study conditions captured.}
      \label{fig:dota_examples_a}
  \end{subfigure} \\[1mm]
  \begin{subfigure}{\textwidth}
      \includegraphics[width=\textwidth]{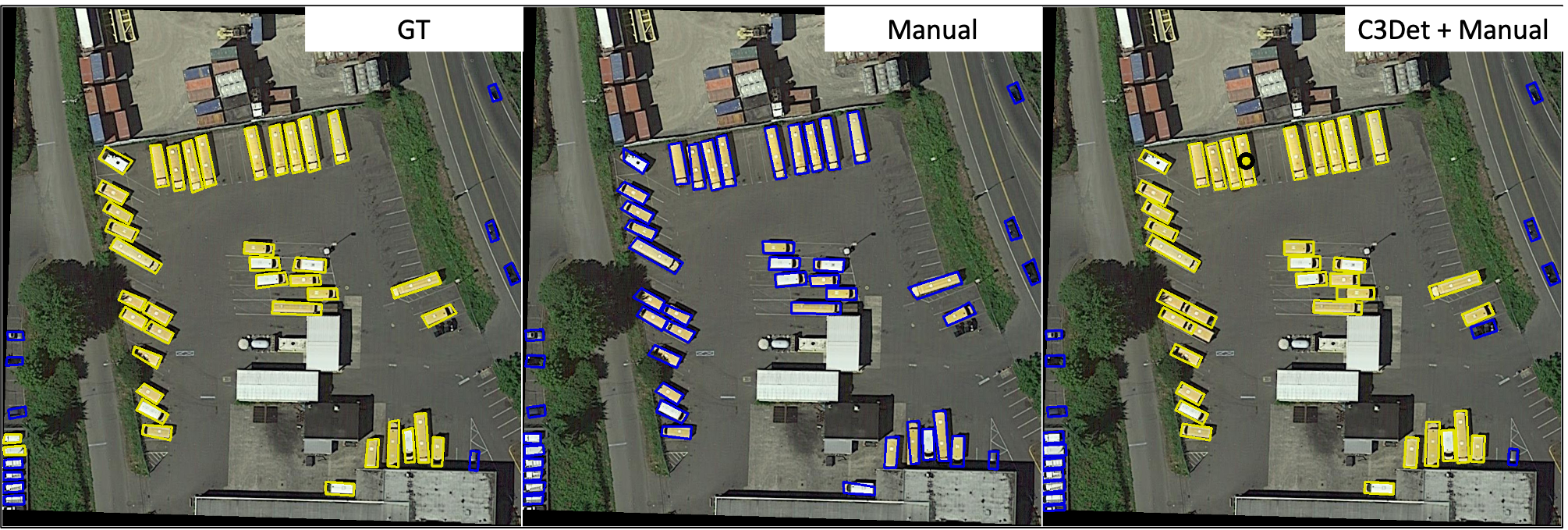}
      \caption{Our novice annotators can make critical mistakes (mislabeling \emph{large-vehicle} objects as \emph{small-vehicle}), which can be corrected by \netname.}
      \label{fig:dota_examples_b}
  \end{subfigure} \\[1mm]
  \begin{subfigure}{\textwidth}
      \includegraphics[width=\textwidth]{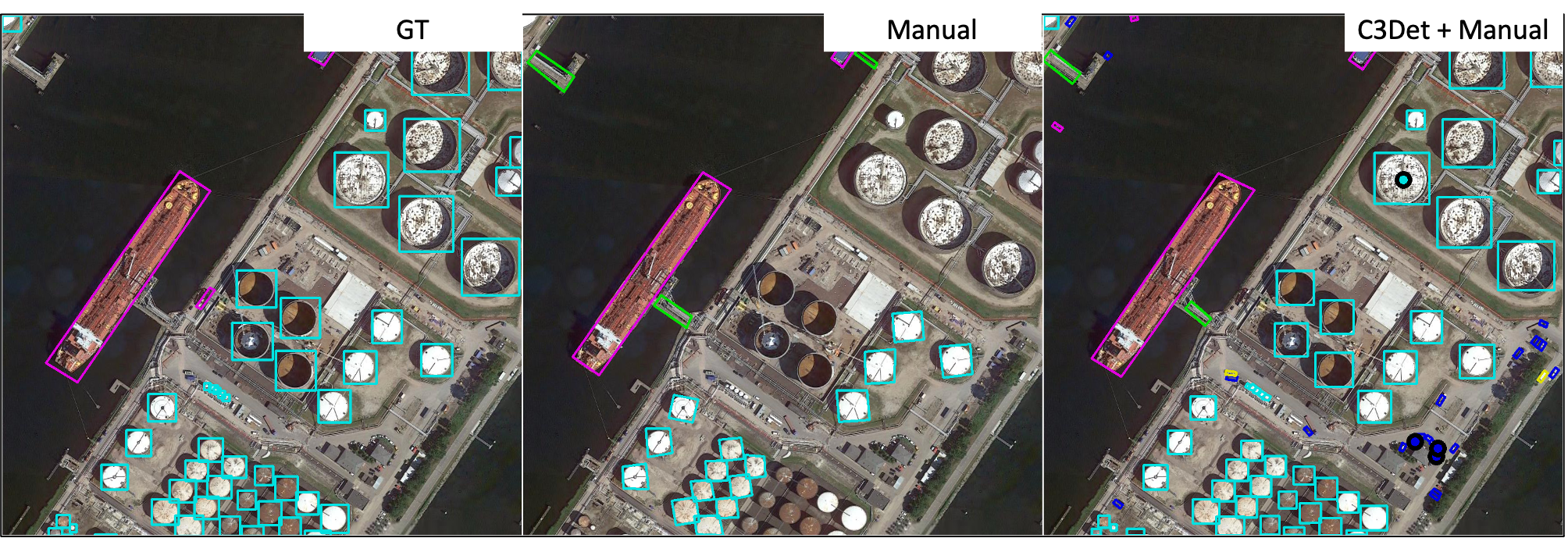} 
      \caption{\netname can allow for better completion in the case where annotators are unsure about certain objects, or do not sufficiently zoom in to annotate very tiny objects.}
      \label{fig:dota_examples_c}
  \end{subfigure} \\[1mm]
  \caption{
  \textbf{Samples from the User Study.}
  Example images of \smalldota with ground-truth, manual, and \netname + manual with annotations.
  The boxes and dots represents final annotated objects and user inputs from real annotators.
  To visualize the user inputs, we draw larger dots compared to the actual User Study GUI.
  \label{fig:dota_examples}
  }
\end{figure*}

\end{document}